\documentclass[12pt]{clear2027}

\jmlrproceedings{Preprint}{Preprint}
\jmlrvolume{}
\jmlrpages{}
\jmlryear{2026}

\usepackage{iftex}
\ifXeTeX
  \usepackage{fontspec}
\else
  \usepackage[T1]{fontenc}
  \usepackage{times}
\fi

\usepackage{booktabs}
\usepackage{mathtools}
\usepackage{xspace}
\usepackage{multirow}
\usepackage{array}
\usepackage[capitalize,noabbrev]{cleveref}

\newcommand{\tbd}[1][]{\textcolor{blue!55!black}{[\,\ifx\relax#1\relax\textbullet\else#1\fi\,]}\xspace}

\newcommand{\figwidth}[1]{\dimexpr #1\textwidth*85/100\relax}
\newcommand{\sidewidth}[1]{#1\linewidth}
\newcommand{\sidesep}{\hfill}

\newcommand{\nPairsTotal}{781\xspace} 
\newcommand{\nScored}{651\xspace} 
\newcommand{\nDirect}{222\xspace} 
\newcommand{\nNotDirect}{429\xspace} 
\newcommand{\nUncertain}{130\xspace} 
\newcommand{\nDirPersistent}{140\xspace} 
\newcommand{\nDirOneTime}{62\xspace} 
\newcommand{\nDirNoShape}{20\xspace} 
\newcommand{\nOciDirect}{2\xspace} 
\newcommand{\nOciScored}{66\xspace} 
\newcommand{\aurocTargetPrior}{0.860\xspace} 
\newcommand{\nFiveOhTwoDirect}{2\xspace} 
\newcommand{\nFiveOhTwoScored}{50\xspace} 
\newcommand{\nStateStateDirect}{4\xspace} 
\newcommand{\nStateStatePairs}{121\xspace} 
\newcommand{\nActions}{71\xspace} 
\newcommand{\nItems}{10\xspace} 
\newcommand{\actFiringsMedian}{620\xspace} 
\newcommand{\actFiringsMin}{24\xspace} 
\newcommand{\actFiringsMax}{20,826\xspace} 
\newcommand{\actFiringRateMedian}{0.7\%\xspace} 
\newcommand{\nActionsUnderHundred}{10\xspace} 
\newcommand{\actTopRate}{23\%\xspace} 
\newcommand{\actTopItem}{5.02\xspace} 
\newcommand{\nFirms}{2,901\xspace} 
\newcommand{\nFirmQuarters}{89,931\xspace} 
\newcommand{\stateCompleteness}{98.2\%\xspace} 
\newcommand{\nQuarters}{31\xspace} 
\newcommand{\nStates}{11\xspace} 
\newcommand{\aurocObs}{0.658\xspace} 
\newcommand{\aurocPooled}{0.609\xspace} 
\newcommand{\nActionsUsable}{43\xspace} 
\newcommand{\nActionsTotal}{70\xspace} 
\newcommand{\nActionsExcluded}{27\xspace} 
\newcommand{\nComparisons}{869\xspace} 
\newcommand{\aurocAggMin}{0.638\xspace} 
\newcommand{\aurocAggMax}{0.658\xspace} 
\newcommand{\nullLabelPerm}{0.501\xspace} 
\newcommand{\excessShiftFirm}{0.120\xspace} 
\newcommand{\nullShiftFirmUnr}{0.556\xspace} 
\newcommand{\nullReps}{400\xspace} 
\newcommand{\pFloor}{0.0025\xspace} 
\newcommand{\nNullKinds}{8\xspace} 
\newcommand{\excessMin}{0.102\xspace} 
\newcommand{\excessMax}{0.157\xspace} 
\newcommand{\excessRatio}{1.5\xspace} 
\newcommand{\shiftDistOneMax}{0.675\xspace} 
\newcommand{\shiftDistOneMin}{0.627\xspace} 
\newcommand{\shiftAdmLow}{0.480\xspace} 
\newcommand{\shiftAdmHigh}{0.507\xspace} 
\newcommand{\shiftDistFifteen}{0.554\xspace} 
\newcommand{\zeroDirectFpr}{0.273\xspace} 
\newcommand{\zeroDirectFprNull}{0.083\xspace} 
\newcommand{\zeroNotDirectRecall}{0.611\xspace} 
\newcommand{\zeroNotDirectRecallNull}{0.053\xspace} 
\newcommand{\aurocReliable}{0.662\xspace} 
\newcommand{\excessReliable}{0.120\xspace} 
\newcommand{\placeboAuroc}{0.568\xspace} 
\newcommand{\placeboNull}{0.540\xspace} 
\newcommand{\placeboP}{0.15\xspace} 
\newcommand{\placeboReps}{100\xspace} 
\newcommand{\nZeroDirect}{25\xspace} 
\newcommand{\nZeroNotDirect}{2\xspace} 
\newcommand{\sizeRuleMin}{0.034\xspace} 
\newcommand{\sizeRuleMax}{0.056\xspace} 
\newcommand{\sizeNaiveTen}{0.074\xspace} 
\newcommand{\sizeWaldTen}{0.156\xspace} 
\newcommand{\sizeWaldEighty}{0.062\xspace} 
\newcommand{\sizeReps}{500\xspace} 
\newcommand{\nWaldUnreliable}{87\xspace} 
\newcommand{\aurocObsWithinTarget}{0.564\xspace} 
\newcommand{\nStatesPersistent}{5\xspace} 
\newcommand{\rhoHighMin}{0.83\xspace} 
\newcommand{\rhoHighMax}{0.93\xspace} 
\newcommand{\rhoLowMin}{-0.05\xspace} 
\newcommand{\rhoLowMax}{0.43\xspace} 
\newcommand{\rhoKendallAboveOneMax}{64\%\xspace} 
\newcommand{\rhoKendallAboveOneMin}{42\%\xspace} 
\newcommand{\rhoRevenue}{0.83\xspace} 
\newcommand{\iccMin}{0.018\xspace} 
\newcommand{\iccMax}{0.795\xspace} 
\newcommand{\effNMin}{3,645\xspace} 
\newcommand{\effNMax}{58,452\xspace} 
\newcommand{\colContMax}{0.83\xspace} 
\newcommand{\nIdentNotComputable}{3\xspace} 
\newcommand{\cleanSurplusUnderPct}{56\%\xspace} 
\newcommand{\opNetUnderPct}{69\%\xspace} 
\newcommand{\balNegPct}{0.7\%\xspace} 
\newcommand{\detDirQLow}{31\%\xspace} 
\newcommand{\detDirQHigh}{80\%\xspace} 
\newcommand{\nDirUnderReq}{132\xspace} 
\newcommand{\detDirUnderReq}{37\%\xspace} 
\newcommand{\detDirOverReq}{69\%\xspace} 
\newcommand{\detDirCalled}{50\%\xspace} 
\newcommand{\nUnanimous}{442\xspace} 
\newcommand{\nUnanimousNot}{347\xspace} 
\newcommand{\nUnanimousDir}{95\xspace} 
\newcommand{\detDirTrim}{55\%\xspace} 
\newcommand{\naiveFlipShare}{16\%\xspace} 
\newcommand{\aurocMinPNaive}{0.587\xspace} 
\newcommand{\detDirNaive}{45\%\xspace} 
\newcommand{\detNotNaive}{40\%\xspace} 
\newcommand{\aurocMinPClustered}{0.640\xspace} 
\newcommand{\detDirClustered}{50\%\xspace} 
\newcommand{\detNotClustered}{37\%\xspace} 
\newcommand{\kappaFleiss}{0.60\xspace} 
\newcommand{\nKappaPairs}{780\xspace} 
\newcommand{\nDisagree}{281\xspace} 
\newcommand{\powFpr}{0.064\xspace} 
\newcommand{\powFiveTwoK}{0.65\xspace} 
\newcommand{\powTenTwoHundred}{0.47\xspace} 
\newcommand{\powReps}{200\xspace} 
\newcommand{\nRequiredTen}{752\xspace} 
\newcommand{\nUnderTen}{40\xspace} 
\newcommand{\nUnderTotal}{71\xspace} 
\newcommand{\suppDirMedian}{528\xspace} 
\newcommand{\suppNotMedian}{702\xspace} 
\newcommand{\suppTopQNotDirect}{83\%\xspace} 
\newcommand{\suppOtherQNotDirectMax}{63\%\xspace} 
\newcommand{\bundleMedian}{89\%\xspace} 
\newcommand{\bundleMin}{62\%\xspace} 
\newcommand{\nBundleSameItem}{18\xspace} 
\newcommand{\nBundleActions}{71\xspace} 
\newcommand{\kurtRawMin}{4,043\xspace} 
\newcommand{\kurtRawMax}{83,688\xspace} 
\newcommand{\kurtWinsMin}{18\xspace} 
\newcommand{\kurtWinsMax}{451\xspace} 
\newcommand{\topSsWinsMin}{27\%\xspace} 
\newcommand{\topSsWinsMax}{76\%\xspace} 
\newcommand{\rTwoAuxTerc}{0.0082\xspace} 
\newcommand{\rTwoAuxLog}{0.0180\xspace} 
\newcommand{\bpClassicMedian}{66,896\xspace} 
\newcommand{\bpKoenkerMedian}{633\xspace} 
\newcommand{\bpClusterMedian}{70\xspace} 
\newcommand{\bpCells}{2,604\xspace} 
\newcommand{\shiftDistFour}{0.517\xspace} 
\newcommand{\shiftDistTwo}{0.629\xspace} 

\newcommand{\sensUnanAuroc}{0.647\xspace} 
\newcommand{\sensUnanWT}{0.516\xspace} 
\newcommand{\sensNoOppAuroc}{0.687\xspace} 
\newcommand{\sensStrictAuroc}{0.566\xspace} 
\newcommand{\sensUncDirAuroc}{0.643\xspace} 
\newcommand{\sensUncVoteAuroc}{0.658\xspace} 
\newcommand{\sensUnanPairs}{442\xspace} 
\newcommand{\sensContestedPairs}{57\xspace} 
\newcommand{\sensContestedComparisons}{14\xspace} 
\newcommand{\sensAnnotMin}{0.649\xspace} 
\newcommand{\sensAnnotMax}{0.668\xspace} 
\newcommand{\sensStrictRelabelled}{127\xspace} 
\newcommand{\sensUnanExcess}{0.098\xspace} 
\newcommand{\sensUnanP}{0.007\xspace} 
\newcommand{\sensNullQ}{0.581\xspace} 
\newcommand{\sensFlipNotToDir}{30\%\xspace} 
\newcommand{\sensFlipBoth}{20\%\xspace} 
\newcommand{\sensFlipMaxRate}{50\%\xspace} 
\newcommand{\sensAdvFlips}{12\xspace} 
\newcommand{\sensAdvShare}{1.8\%\xspace} 
\newcommand{\sensAdvFlipsNonUnan}{14\xspace} 
\newcommand{\sensAdvDsPool}{17\xspace} 
\newcommand{\sensAdvDsAuroc}{0.641\xspace} 
\newcommand{\sensAdvDsUPool}{10\xspace} 
\newcommand{\sensAdvDsUAuroc}{0.639\xspace} 
\newcommand{\sensDsMean}{0.665\xspace} 
\newcommand{\sensDsLow}{0.654\xspace} 
\newcommand{\sensDsUMean}{0.671\xspace} 
\newcommand{\sensDsULow}{0.660\xspace} 
\newcommand{\sensDsAgree}{98.2\%\xspace} 
\newcommand{\sensDsUAgree}{96.6\%\xspace} 
\newcommand{\sensDetNotN}{148\xspace} 
\newcommand{\sensDetNotVoted}{11\xspace} 
\newcommand{\sensDetNotDs}{5.7\xspace} 
\newcommand{\sensDetNotDsU}{12.7\xspace} 
\newcommand{\sensTwFirmAuroc}{0.640\xspace} 
\newcommand{\sensTwFqAuroc}{0.643\xspace} 
\newcommand{\sensTwFirmWT}{0.569\xspace} 
\newcommand{\sensTwFqWT}{0.599\xspace} 
\newcommand{\sensTwFirmDetDir}{50\%\xspace} 
\newcommand{\sensTwFqDetDir}{43\%\xspace} 
\newcommand{\sensTwFirmDetNot}{37\%\xspace} 
\newcommand{\sensTwFqDetNot}{24\%\xspace} 
\newcommand{\sensTwSeRatio}{1.07\xspace} 
\newcommand{\sensPreScaleAuroc}{0.625\xspace} 
\newcommand{\sensPreScaleWT}{0.571\xspace} 
\newcommand{\sensRvDir}{0.013\xspace} 
\newcommand{\sensKSizeDir}{16\xspace} 
\newcommand{\sensKActDir}{42\xspace} 
\newcommand{\sensKSizeNot}{11\xspace} 
\newcommand{\sensKActNot}{38\xspace} 
\newcommand{\sensKQuarterNot}{33\xspace} 

\newcommand{\E}{\mathbb{E}}
\newcommand{\caufin}{CauFin\xspace}

\title[Where Causal Discovery Breaks on Real Data]{Where Causal Discovery Breaks on Real Data:\\
  Measuring the Assumptions One Property at a Time}

\clearauthor{%
  \Name{Padmaja Jonnalagedda\textsuperscript{*\textdagger}}
    \Email{saisri\_jonnalagedda@intuit.com}\\
  \addr Intuit AI Research
  \AND
  \Name{Avinash Baidya\textsuperscript{*\textdagger}}
    \Email{avinash\_baidya@intuit.com}\\
  \addr Intuit AI Research
  \AND
  \Name{Ruocheng Guo\textsuperscript{\textdaggerdbl}}\\
  \addr Microsoft
  \AND
  \Name{Kamalika Das}\\
  \addr Intuit AI Research
}

\begin{document}

\maketitle

\begin{center}
  \small
  \textsuperscript{*}Equal contribution.\quad
  \textsuperscript{\textdagger}Corresponding authors.\quad
  \textsuperscript{\textdaggerdbl}Work done while at Intuit AI Research.
\end{center}

\begin{abstract}
Causal discovery is usually evaluated on simulators or on systems whose ground
truth follows from physics or design. When it fails on other real data, the
failure is attributed to assumptions that may not hold, but which ones fail,
and by how much, is rarely measured. We measure the observable properties of
the data they bear on, one property at a time, on a real panel. The case study follows \nFirms US public firms over \nQuarters quarters,
with \nActions types of corporate action from SEC 8-K filings, \nStates
financial-statement items, and an expert graph over \nScored action--state
pairs. The panel breaks measurable requirements of discovery: tails that survive winsorisation, balance-sheet items and revenue
close to random walks, and interventions too rare to detect effects of a tenth of a standard
deviation in \nUnderTen of \nUnderTotal actions. Even so, a marginal test
ranks expert-labelled direct above not-direct pairs within each action (AUROC \aurocObs)
better than every null we built, so the panel does carry information about
which pairs are direct. Tuned discovery methods, in configurations feasible
at this panel size, recover less of it than a
simple pairwise cross-correlation when pairs are ranked within each action, and
no more when they are ranked within each target, which removes what the target
alone predicts. Most direct pairs (\nDirUnderReq of \nDirect) involve actions
below the support needed for 80\% power against a 0.1-SD effect.
For flow targets, conditioning on the target's other expert-labelled parents, often filed
alongside the action, erodes the marginal discrimination, and
most not-direct associations survive it: even with the graph's parent sets
supplied, the expected separation does not emerge. For each
property, we give a diagnostic computable before any discovery method is
run.
\end{abstract}


\section{Introduction}
\label{sec:intro}

Most of what we know about causal discovery methods comes from data where the
answer is known by construction: simulators, which can reward artefacts of
the generator \citep{reisach2021beware}, and a few real systems whose ground
truth follows from physics or design, such as river networks
\citep{stein2025causalrivers}, climate model output \citep{runge2020c4c} and
industrial processes \citep{mogensen2024industrial}. They share a regime:
continuous variables, observed densely, with mechanisms that are stable over
time. Much of the data people want to run discovery on differs on each count.
Business, clinical and economic panels are often short and wide, with
interventions that are discrete, rare and dated, and outcomes with extreme
tails. When discovery does poorly on such data, the usual explanation is a
list of assumptions that might be violated. They are rarely measured, so a
practitioner cannot tell which violation matters, or whether a better method
would have helped.

This paper measures the observable properties of the data that bear on these
assumptions and on the statistical machinery of discovery, before a method is
run. Causal sufficiency or faithfulness cannot in general be verified from
observational data, and we do not claim to verify them; tails, persistence,
collinearity and support can be measured. We take
one real panel with an expert graph that is not estimated from it and, for each
property, report a diagnostic, its value, the assumption it bears on, and what
it does to discovery and to its evaluation.

The panel covers \nFirms US public firms over \nQuarters quarters. The
interventions are \nActions types of corporate action filed on SEC Form 8-K,
and the outcomes are \nStates financial-statement items. Annotators labelled
each action--state pair as direct, not direct or uncertain, and judged
each direct effect one-time or persistent. The shapes are expert judgements, not estimates,
and no method is asked to recover a lag. Finance is the setting, not the
subject.

The paper follows that order. The panel breaks the assumptions behind
discovery in measurable ways: extreme tails, near-unit-root persistence,
strongly interrelated states, and interventions too rare for small effects
(\cref{sec:properties}). The same properties force a calibrated evaluation:
shift nulls sit above 0.5 and alias the effect at short distances, and
scoring each pair by how often its target is labelled direct for other
actions gives \aurocTargetPrior (\cref{sec:evaluation}). We compute these
references with one marginal test, an event-study comparison of firms that
fire an action with firms that do not. It ranks direct above not-direct pairs within an action at
AUROC \aurocObs, above all \nNullKinds nulls, so the data are not empty, and
its detection rates later show which direct pairs a marginal comparison can
see. Discovery methods from each family, tuned on the labels and scored the
same way, reach 0.528--0.700 within action, below pairwise cross-correlation
(0.824), and none has a larger excess over its own null than the baselines.
Within target, which removes what the target alone predicts, none is above
cross-correlation either (0.643 against at most 0.579; \cref{sec:methods}).
Two questions account for the scores: why so few direct pairs are seen even
marginally, and why conditioning does not separate the rest. Most direct
pairs sit on actions too rare for a marginal comparison to find an effect of
0.1 SD (\nDirUnderReq of the \nDirect), the share detected rises from
\detDirQLow in the lowest support quartile to \detDirQHigh in the highest,
and extreme rows hide part of the rest (\cref{sec:missed}). For flow targets,
conditioning on the expert parents lowers the AUROC from 0.647 to 0.533,
on levels mostly through same-quarter parents, which shrink direct
coefficients but leave not-direct ones intact. Most not-direct detections
survive: the panel carries dependence the graph does not account for. It
survives firm and quarter effects and dropping the filing quarter, annotator
disagreement explains few detections, and a confounder must be \sensKSizeNot
times as strong as log firm size. Stock
targets fail before conditioning: they do not separate even marginally
(\cref{sec:failure}).

We contribute \caufin, a benchmark of corporate actions and financial states
with an expert graph (\cref{sec:benchmark}); a catalogue of data properties
tied to discovery assumptions, with diagnostics that transfer to other panels
(\cref{sec:properties,sec:checklist}); an evaluation protocol calibrated
against matched nulls (\cref{sec:evaluation}); and
discovery methods from each family scored under that protocol, in the
configurations this panel size allows, with an account of why they
fail (\cref{sec:methods,sec:failure}). Because the expert graph is partial
and may contain errors, a low score alone does not show that a method fails;
we bound how much label error could explain (\cref{app:sensitivity}). The
diagnostics apply to any panel of rare, dated interventions; the values we
report describe this one. We will release the panel, the labels, the code and the extraction
prompts upon acceptance.

\section{Related work}
\label{sec:related}

\textbf{Causal discovery benchmarks.}
Benchmarks are synthetic, semi-synthetic or real
\citep{glymour2019review, assaad2022survey}. Synthetic ones generate data from
known graphs \citep{zheng2018dags, lachapelle2020gradient} and can reward
artefacts of the generator \citep{reisach2021beware}; CausalTime
\citep{cheng2024causaltime} learns generators from real series, and TimeGraph
\citep{ferdous2025timegraph} and CausalDynamics
\citep{herdeanu2025causaldynamics} scale generation to thousands of coupled
continuous systems. Real benchmarks with known structure (protein signalling
\citep{sachs2005causal}, cause--effect pairs \citep{mooij2016tuebingen}, rivers
\citep{stein2025causalrivers}, climate models \citep{runge2020c4c}, industrial
processes \citep{mogensen2024industrial}) share one regime: continuous
variables, observed densely, under stable mechanisms, and most ask for the
whole graph. CausalBench \citep{chevalley2023causalbench} pairs discrete
interventions with continuous outcomes, but the interventions are designed;
FinCARE \citep{zuniga2025fincare} takes firm ground truth from 10-K filings
but simulates the data. We know of no benchmark whose interventions occur
naturally, rarely and on known dates in an observational panel, with edges
labelled from domain mechanism rather than from the data.
\citet{gentzel2019case} argue for interventional evaluation; our labels are
expert judgements, not interventions.

\textbf{Methods and their assumptions.}
Methods test conditional independence (PC \citep{spirtes2000causation},
PCMCI+ \citep{runge2019pcmci, runge2020pcmciplus}, LPCMCI
\citep{gerhardus2020lpcmci}), optimise a score over graphs (GES
\citep{chickering2002optimal}, DYNOTEARS \citep{pamfil2020dynotears}), orient
edges with non-Gaussian noise (VAR-LiNGAM \citep{hyvarinen2010varlingam}), ask
whether one variable's past predicts another's \citep{granger1969investigating},
or use language models as priors or graph generators \citep{ban2023causal,
kiciman2023causal, jiralerspong2024efficient, roy2025causalllm}. Strong
faithfulness fails for a large share of parameters \citep{uhler2013geometry},
and single wrong independence decisions propagate through PC
\citep{faltenbacher2025pc}. Extensions relax one assumption at a time: latent
confounding \citep{gerhardus2020lpcmci}, nonstationarity
\citep{huang2020cdnod, gao2023semistationary, balsells2025sdci,
wiecksosa2025single}, hidden confounding across environments
\citep{karlsson2023hidden}. Mixed binary and continuous data strain several
families at once: partial-correlation tests assume a Gaussian model, and
additive noise is ill-defined for a binary cause; MiTCD \citep{chen2025mitcd}
first recovers latent continuous variables from the discrete ones. We know of
no work that measures the full set of requirements on one real dataset
against an expert graph.

\textbf{Evaluation and negative controls.}
Our evaluation follows negative-control calibration in medicine
\citep{schuemie2014interpreting, schuemie2016robust, schuemie2018empirical}
and \citet{bertrand2004much}, who showed that placebo interventions in
serially correlated panels are declared significant far too often unless the
dependence is modelled. In perturbation-response prediction, simple baselines
matched deep models \citep{ahlmanneltze2025deep} until the metrics were
recalibrated against positive and negative controls \citep{miller2025deep}: a
precedent for checking the metric before the method.

\textbf{Causal inference in finance.}
Empirical finance identifies single effects, not graphs, with instruments,
discontinuities and difference-in-differences \citep{angrist2009mostly}. Our
marginal test is close to an event study \citep{mackinlay1997event}; unlike
staggered-adoption estimators \citep{callaway2021difference,sun2021estimating},
it does not correct for effects that differ across adoption cohorts. Its
firm-clustered inference \citep{petersen2009estimating, cameron2015practitioner}
is unreliable when few clusters are treated \citep{mackinnon2018wild}, the
usual case for a rare action.

\section{The \caufin benchmark}
\label{sec:benchmark}
\label{sec:case}

\caufin is a partial discovery benchmark. A method may return any graph over
the panel's variables, but one block is scored: edges from an action filed in
quarter $t$ to a state in the next four quarters.

\begin{figure*}[t]
\centering
\includegraphics[width=\figwidth{0.9}]{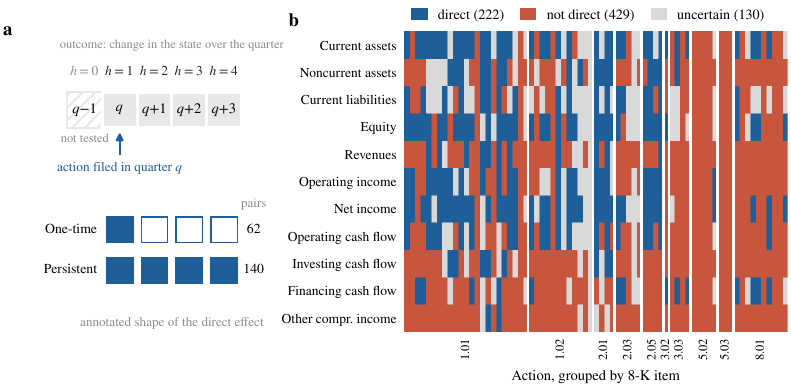}
\caption{The benchmark. (a) States are recorded at the start of each
quarter, so the first change an action filed in quarter $q$ can produce is the
change over $q$, indexed $h=1$; $h=0$, the last change before the filing, is
not tested. Annotators judged each direct effect one-time or persistent
(counts are pairs; cells are schematic; \nDirNoShape direct pairs have no
shape); the shapes are judgements, not estimates. (b) Labels of all
\nPairsTotal action--state pairs, actions grouped by 8-K item. Near-uniform
rows and blocks feed the label-only baseline of \cref{sec:identity}.}
\label{fig:setup}
\end{figure*}

\textbf{Edge labels.}
We observe firms $i = 1, \dots, N$ over quarters $t = 1, \dots, T$, with binary
actions $a^{k}_{i,t}$ and continuous states $s^{j}_{i,t}$. For horizon
$h \in \{1,\dots,4\}$ let $W_{t,h}$ be every other observed variable dated
before $t+h$, including the target's own past. $W_{t,h}$ includes variables
dated after $t$ that the action may itself move, so following
\citet{pearl2009causality} we use the controlled direct effect, with all of
them held fixed:
\begin{equation}
  \begin{split}
\label{eq:direct}
\Delta^{kj}_h(w) = \E\bigl[s^{j}_{t+h} \mid \mathrm{do}(a^{k}_{t}=1,\, W_{t,h}=w)\bigr] 
 \\ - \E\bigl[s^{j}_{t+h} \mid \mathrm{do}(a^{k}_{t}=0,\, W_{t,h}=w)\bigr].
  \end{split}
\end{equation}
The pair is \emph{direct} if $\Delta^{kj}_h(w) \neq 0$ for some $h$ and $w$:
the action moves the state by a route that passes through no other observed
variable. It is \emph{not direct} if $\Delta^{kj}_h(w) = 0$ for every $h$ and
$w$: any effect runs through other observed variables, or there is none. A pair is \emph{uncertain} if the
answer depends on an attribute $u$ of the event that the panel does not
record, so that $\Delta^{kj}_h(w;u) \neq 0$ for some $u$ and $= 0$ for others.
The attribute is either a detail of the action (an exit plan moves other
comprehensive income only if it involves a foreign subsidiary or a pension
curtailment) or a later management decision (an at-the-market equity
programme moves equity only if it is used). Uncertain pairs are left out of
every metric. The definitions are relative to the variable set: a route
through an unrecorded quantity (working capital, headcount, a subsidiary's
balance sheet) counts as direct, and the labels are judged in this sense.

\textbf{Panel.}
\label{sec:panel}
The panel covers \nFirms firms, identified by SEC Central Index Key, over the
\nQuarters quarters from 2018Q2 to 2025Q4 (\nFirmQuarters firm-quarters,
\cref{tab:stats}). It keeps only firms present in every quarter, so firms
that listed, delisted or were acquired are absent; 1.8\% of
state values are unreported.

\textbf{Actions.}
The \nActions action types are subtypes of \nItems 8-K items
(\cref{tab:taxonomy}), kept because they record voluntary corporate decisions
and fire often enough to test; subtypes separate events under one item whose
accounting differs. A language model assigns the events each 8-K discloses to
subtypes (3\% by embedding similarity instead), and events that fit no
subtype are dropped. An action is binary per firm-quarter and dated by the
calendar quarter of its filing. The median action fires \actFiringsMedian times
(\actFiringRateMedian of firm-quarters), the rarest \actFiringsMin times, and
one item-\actTopItem subtype in \actTopRate of all firm-quarters.
An audit of 30 events finds none invented and two with the wrong subtype,
and about 1.6\% of firm-quarters miss an event; a missed or invented event
hits an action's direct and not-direct targets alike and pulls the
within-action AUROC towards 0.5, while a wrong subtype has no fixed sign
(\cref{app:data}).

\textbf{States.}
The \nStates states are line items from the firms' 10-Q and 10-K filings:
current and noncurrent assets, current liabilities, equity, revenues,
operating income, net income, the three cash-flow totals and other
comprehensive income. There are no totals for assets, liabilities or the
change in cash. Each is dated by the calendar quarter in which the reporting
period ends. Flows filed year to date are converted to single quarters,
and an audit of 40 values against the XBRL filings finds none of the 31
it could verify wrong (\cref{app:data}). \stateCompleteness of values are reported; the rest are left
missing, not imputed, and drop out of any regression that uses them; exact
zeros, some probably unreported, are kept as reported (\cref{app:data}).
Each state is moved to the start of the quarter (\cref{fig:setup}a), divided
by the firm's median total assets (current plus noncurrent, so current assets
alone where noncurrent assets is zero, and the median across firms for the
two firms whose median is zero), and winsorised at the 1st and 99th
percentiles within state and quarter.

\textbf{Annotation.}
\label{sec:graph}
Three annotators labelled each action--state pair as direct, not direct or
uncertain, judging whether an accounting mechanism exists rather than how
large it is (\cref{app:labels}). \nDirect pairs are direct, \nNotDirect not
direct and \nUncertain uncertain, leaving \nScored scored pairs.
The \nStateStatePairs state--state pairs were also labelled, at lags 1--4, but
are not scored; the \nStateStateDirect direct ones are all self-loops of the four
stocks, so no flow has a self-edge and no state is a lagged direct parent of
another; identities that hold within a quarter, such as revenue entering net
income, are not lagged effects.
The annotators followed a written rubric without seeing the panel data,
and agree unanimously on \nUnanimous scored pairs but on only \nUnanimousDir of the \nDirect
direct ones (Fleiss' $\kappa$ \kappaFleiss; \cref{app:labels}). One item-1.02
subtype has all its pairs uncertain, so \nActionsTotal of the \nActions
actions enter the metrics. The labels are structured by row and block
(\cref{fig:setup}b): other comprehensive income is direct in \nOciDirect of
\nOciScored pairs, personnel changes (item 5.02) in \nFiveOhTwoDirect of
\nFiveOhTwoScored. The labels encode the plausibility of an accounting
mechanism, not an experimentally verified effect.

\textbf{Timing.}
Labels belong to the pair, not to a horizon. The annotators also judged each
direct effect one-time (a change in a single quarter) or persistent (carrying
into later quarters): \nDirPersistent are persistent, \nDirOneTime one-time,
and \nDirNoShape have no shape. The shapes are part of the benchmark but are
not evaluated here; scoring them is left to future work. They fix only the
test window, $h \le 4$; no lag is scored, because the same action takes effect at
different horizons across firms with deal timing, fiscal calendars and
management discretion. States are stamped at the start of the quarter, so for
an action filed in quarter $t$ the state $s_t$ precedes the filing and the
change at horizon $h$ is over quarter $t+h-1$ (\cref{fig:setup}a); every
change the score uses is dated after $s_t$.

\section{Properties of the data}
\label{sec:properties}

For each property we give a diagnostic and its value on the panel;
\cref{sec:failure} takes up what the properties hide from a
marginal comparison and predict for discovery methods.

\begin{figure*}[t]
\begin{minipage}[c]{0.53\linewidth}
\centering
\includegraphics[width=\linewidth]{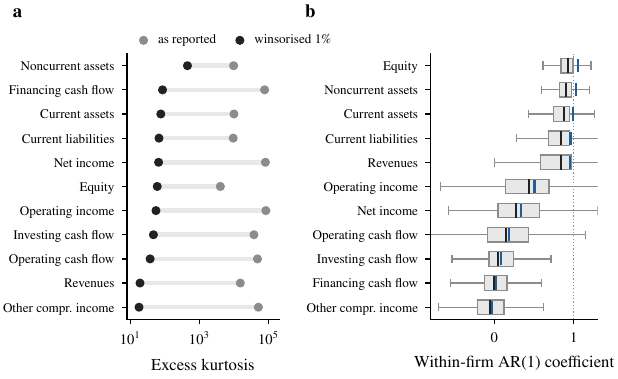}
\end{minipage}\hfill
\begin{minipage}[c]{0.44\linewidth}
\caption{(a) Excess kurtosis of each scaled level, as reported and after 1\%
winsorisation (log scale). (b) Within-firm AR(1) coefficients of the levels;
boxes span firms, the blue tick is the median after Kendall's correction.}
\label{fig:variables}
\end{minipage}
\end{figure*}

\subsection{Heavy tails}
\label{sec:tails}

The scaled levels have excess kurtosis of \kurtRawMin--\kurtRawMax as
reported and \kurtWinsMin--\kurtWinsMax after winsorising
(\cref{fig:variables}a). The top 1\% of rows still carry \topSsWinsMin to
\topSsWinsMax of each state's sum of squares, so for the most tail-heavy
states a least-squares fit on \nFirmQuarters rows is dominated by fewer than a
thousand.

\subsection{Persistence}
\label{sec:persistence}

Persistence splits the states in two, though not into stocks and flows. The
median within-firm AR(1)
coefficient is \rhoHighMin--\rhoHighMax for \nStatesPersistent states, the
four balance-sheet items and revenue, and $\rhoLowMin$--\rhoLowMax for the
rest (\cref{fig:variables}b); revenue is a flow that behaves like a stock
(\rhoRevenue). Over \nQuarters quarters the estimate is biased towards zero by
roughly $(1+3\rho)/T$ \citep{kendall1954note}; after that correction,
\rhoKendallAboveOneMin--\rhoKendallAboveOneMax of
firms have a coefficient above one for the persistent states: over \nQuarters
quarters they are hard to tell from random walks. The residuals of a level
regression on a rare binary action then inherit that persistence, and
the firm-quarters are worth only \effNMin--\effNMax independent observations
(ICC \iccMin--\iccMax). The one-quarter change is close to stationary.
Persistence also decides which shift nulls are valid (\cref{sec:shift}).

Heteroscedasticity against firm size is real but small: a Breusch--Pagan
test rejects in all \bpCells cells even with cluster-robust inference, but
the auxiliary $R^2$, which we report instead, has median \rTwoAuxLog
(\cref{app:extra}).

\subsection{Collinearity}
\label{sec:degeneracy}

Accounting ties line items together, and deterministic relations violate
faithfulness outright. Which ones can be checked depends on what is reported:
here \nIdentNotComputable standard identities cannot be computed at all
(\cref{app:extra}). No state is determined by the others, but some are
largely explained by them: regressed on the other ten at the same date, the
best-explained state has $R^2$ \colContMax, which leaves a test that
conditions on the others less unique variation.

\subsection{Sparse and bundled interventions}
\label{sec:sparse}

\nActionsUnderHundred of the \nActions actions fire fewer than 100 times. To see
what support buys we plant an additive effect of $\delta$ standard deviations
into the real panel at random firm-quarters and rerun the marginal test of
\cref{sec:protocol} cell by cell (\cref{app:power}). With no effect planted,
the harness rejects in \powFpr of replicates at nominal 0.05 (the test's own
size is in \cref{app:inference}). An effect of 0.1 SD is found in
\powTenTwoHundred of replicates at 200 firings and needs about \nRequiredTen
for 80\% power, which \nUnderTen of the \nUnderTotal actions lack; at 0.05 SD,
power is only \powFiveTwoK at 2,000 firings. Real actions cluster in fewer
firms than planted ones, and under clustered inference the treated firms set
power, so the grid bounds what the test can see from above; it does not
estimate how much of the graph is visible. Support is also confounded with
the labels (\cref{sec:pooled}).
\label{sec:bundling}
Actions also arrive together: the median action shares its firm-quarter with
another in \bundleMedian of firings, the least bundled in \bundleMin, and for
\nBundleSameItem of the \nBundleActions actions the most frequent partner is a
subtype of the same 8-K item. Effects of actions that always co-occur cannot be
separated. Same-item partners may also be one event that extraction split
in two.

\section{Experimental setup}
\label{sec:setup}

\textbf{Methods.}
We run discovery methods from four families. Constraint-based: PCMCI+
\citep{runge2020pcmciplus}, which assumes causal sufficiency, and SVAR-RFCI
\citep{malinsky2018causal}, which allows latent confounders;
conditional-independence methods are too expensive at this panel size for
more representatives. Continuous optimisation: DYNOTEARS
\citep{pamfil2020dynotears} and its nonparametric variant NTS-NOTEARS
\citep{sun2023ntsnotears}. Functional: longitudinal LiNGAM
\citep{kadowaki2013longitudinal}, one non-Gaussian model per quarter across
firms. Granger: a linear VAR Granger test \citep{granger1969investigating},
fitted on lagged rows that never cross two firms, and MiTCD
\citep{chen2025mitcd}, a neural Granger model that embeds binary variables
before learning a sparse graph. Two pairwise baselines set the floor a method
should clear: lagged cross-correlation averaged over firms, and the
event-triggered average, the mean change after an action with no control
group.

\textbf{Input and tuning.}
Methods receive the quarterly states and binary action indicators with a lag
window of one to four quarters, and search four panel strategies (pooled
levels, firm effects, firm and quarter effects, first differences) crossed
with two state scales (raw dollars, or division by the firm's median total
assets), all winsorised at 1\% and standardised. The transformations apply
only to the states, so actions stay binary and the demeaned arms are not full
within estimators. VAR Granger, longitudinal LiNGAM and the two baselines
have no other hyperparameters and are evaluated under all eight combinations;
DYNOTEARS, NTS-NOTEARS and MiTCD also search regularisation and optimisation
hyperparameters within a common budget of 40 configurations. PCMCI+ and
SVAR-RFCI are too expensive to tune: each runs in one fixed configuration,
with conditioning sets capped at two variables,\footnote{In fixed-preprocessing
sensitivity tests, raising the caps does not improve either method's scores
(\cref{app:discovery-protocol}).} under each preprocessing
combination, and its significance level is swept over four values to rank
edges (\cref{app:methods}); their results show what a feasible instantiation
recovers, not that the unrestricted algorithms fail. Configurations are
selected by pair-level within-action AUROC on the full label set, so every
entry in \cref{tab:methods} is the best over the configurations searched: an
upper bound, not a held-out estimate.

\textbf{Contemporaneous edges.}
Because $s_t$ is fixed before any action in quarter $t$ is filed
(\cref{sec:benchmark}), a contemporaneous edge $a_t \to s_t$ cannot be causal,
and methods are scored on lags one to four only. On quarter-end data the same
first change would sit at lag zero, where a lagged method never looks.

\textbf{Scoring.}
A method's output is collapsed to one score per action--state pair over lags
one to four and scored as in \cref{sec:evaluation}: the within-action AUROC,
its excess over the method's own restricted shift null, and the within-target
AUROC.

\section{How to evaluate in this setting}
\label{sec:evaluation}

On this panel chance is not 0.5. Support differs across actions, the labels
have structure of their own, and a null that keeps the data's properties
scores well above 0.5, so a method at 0.6 can be reading nothing. We fix the
metric and its references before any method is run; most are negative
controls \citep{schuemie2014interpreting}, built to share the data's
properties but not the effect.

\textbf{A marginal test.}
\label{sec:protocol}
Every reference below (the nulls, the placebo, the shift curve) is recomputed
thousands of times through one fixed score; a discovery method is too slow
for that, and harder to read under a null. Rare, dated, binary interventions in
a panel are the setting of the event study \citep{mackinlay1997event}, so we
use the simplest test of that kind: firms that fire an action against firms
that do not, the pooled, one-quarter-change counterpart of the event-triggered
average of \cref{sec:setup} with a control group. The test is also the paper's
instrument: its known size gives the
references in this section, the power of \cref{sec:sparse} and the detection
rates of \cref{sec:missed}. It uses changes because the persistent states make
a level regression on a rare action unreliable (\cref{sec:persistence}). For
action $k$, state $j$ and $h \in \{1,\dots,4\}$ we regress
$s^{j}_{i,t+h} - s^{j}_{i,t+h-1}$ on $a^{k}_{i,t}$, pooled over the
firm-quarters where the change is observed, with firm-clustered errors and
degrees of freedom set by the number of firms that do and do not fire the
action (\cref{app:inference}). A pair's score is $-\log_{10} p$ of a joint
test that any horizon differs from zero; other ways of combining the horizons
give \aurocAggMin to \aurocAggMax (\cref{app:aggregation}). The test is
oversized when few firms fire the action; we flag the \nWaldUnreliable such
pairs.

\textbf{Pooled AUROC ranks by firing frequency.}
\label{sec:pooled}
Direct pairs sit on rarer actions (median \suppDirMedian firings against
\suppNotMedian), and \suppTopQNotDirect of pairs in the top support quartile
are not direct, against at most \suppOtherQNotDirectMax in the others; the
most frequent action, a personnel change (item 5.02), is almost never linked
directly to a line item. Pooled across actions, any score that grows with
sample size partly ranks by firing frequency: the marginal test scores
\aurocPooled pooled and \aurocObs within action.

\textbf{Within-action AUROC.}
We therefore compute the AUROC \emph{within action}, comparing each direct
target of an action with each not-direct target of the same action. All
targets of an action share its firings, so support differences cancel.
\nActionsUsable of \nActionsTotal actions have both classes
(\nComparisons comparisons).
\label{sec:coverage}
The other \nActionsExcluded are invisible to the AUROC: \nZeroDirect have no
direct target and \nZeroNotDirect no not-direct one, so a ranking metric says
nothing about them, and for these we report detection rates instead
(\cref{app:coverage}).

\textbf{What the target alone predicts.}
\label{sec:identity}
The within-action AUROC removes differences between actions but not between
targets, which differ widely in how often they are labelled direct. Scoring
each pair by the share of other actions labelled direct on its target, with no
panel data, already gives \aurocTargetPrior. This is not a method to beat but
a measure of what the target alone supplies, and it can be discounted two
ways. The nulls of \cref{sec:chance} keep labels and
targets fixed, so the excess over a null already leaves it out. Ranking within
target (each score replaced by its percentile among the scored pairs on its
target) removes it from the AUROC itself, so a target-only score gets 0.5, but
also discards real signal shared across a target's actions. Here the two
disagree: the marginal test keeps an excess of \excessShiftFirm over its null
but scores \aurocObsWithinTarget within target, which points to real effects
shared across a target's actions. The within-action AUROC is therefore
descriptive; the excess over the null and the within-target AUROC, which
discount the target, are the evaluation, and we report both throughout.

\begin{figure*}[t]
\centering
\includegraphics[width=\figwidth{1}]{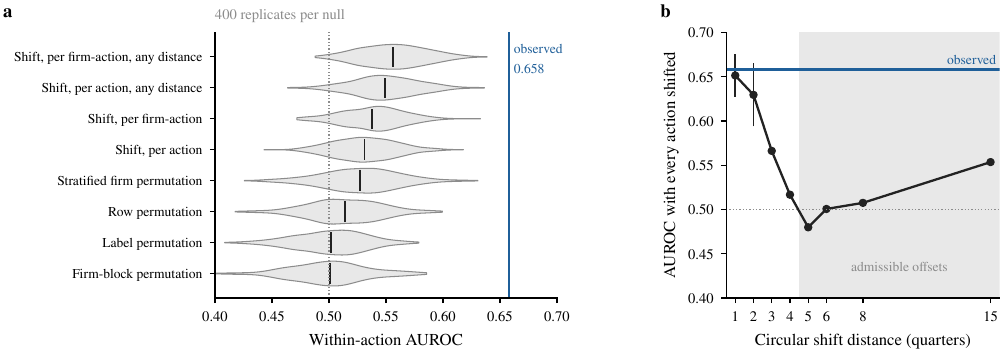}
\caption{(a) Within-action AUROC under each null (\nullReps replicates); the
line is the observed value. (b) AUROC with every action circularly shifted by
the same distance; at one and two the bar spans both directions. Restricted
nulls draw only from the shaded offsets.}
\label{fig:nulls}
\end{figure*}

\textbf{Chance depends on the null.}
\label{sec:chance}
We build \nNullKinds nulls of \nullReps replicates each: label permutation
within action, firm-block permutation with and without activity strata, row
permutation, and circular shifts within firm, per action or per firm, at any
distance or at least five quarters (\cref{app:nulls}). Label and firm-block
permutation centre on \nullLabelPerm and the shifts reach \nullShiftFirmUnr
(\cref{fig:nulls}a, \cref{tab:nulls}), so the excess shrinks from \excessMax
to \excessMin, a factor of \excessRatio, as the null keeps more of the data. A
shift keeps which firms file each action and moves only when. The gap is
composition: filers differ from non-filers more in some states than others,
and direct labels concentrate on particular targets, so even a shifted action
ranks direct pairs above not-direct ones more often than not. The shift null
is the right reference; label and firm-block permutation are the most lenient.
A placebo that scores each action against the changes at $h=-4,\dots,-1$,
four quarters that end before the quarter preceding the filing, gives
\placeboAuroc against \placeboNull under the restricted shift null
(\cref{app:nulls}), an excess about a quarter of the forward one and no
pre-action difference beyond what the null carries.

\textbf{Short shifts are not nulls.}
\label{sec:shift}
The test looks four quarters past each firing, so a shift of one to four
quarters can leave the real effect inside the window: at one quarter the
AUROC is still \shiftDistOneMin--\shiftDistOneMax against \aurocObs observed,
and it falls to \shiftAdmLow--\shiftAdmHigh only at five to eight
(\cref{fig:nulls}b). The primary null therefore draws only shifts of five
quarters or more. Each method's excess in \cref{tab:methods} is against the
same null refitted through the method: its best configuration is rerun on
panels with every action shifted within firm by five quarters or more. Methods
search levels and first differences (\cref{sec:setup}); in levels a persistent
step stays put wherever the shifted indicator lands, so the null keeps part of
any persistent effect and the excess of a levels configuration is
conservative.

\textbf{The panel carries information about the labels.}
\label{sec:headline}
Against these references the marginal test scores \aurocObs. No replicate of
any null reaches it ($p \le \pFloor$), and the excess over the primary null,
the restricted per-firm shift, is \excessShiftFirm. On pairs whose action
fires in at least 80 firms, where the test holds its size in simulation
(\cref{app:inference}), the AUROC is \aurocReliable, an excess of
\excessReliable, so the flagged pairs do not drive the headline. Nor do
contested labels: unanimous pairs alone give \sensUnanAuroc, and random flips
bring the AUROC down to the null's 95th percentile only once
\sensFlipNotToDir of not-direct labels are made direct
(\cref{app:sensitivity}). The panel carries information about which pairs the
annotators call direct. \Cref{sec:methods} scores the methods against the same
kind of reference, each against a null refitted through its own pipeline, and
\cref{sec:missed} asks which direct pairs the marginal test misses and why.

\section{How do discovery methods perform on the panel?}
\label{sec:methods}

{\setlength{\emergencystretch}{1.5em}%
No discovery method reaches cross-correlation (\cref{tab:methods}).
The best within action, longitudinal LiNGAM, scores 0.700 with an excess of
0.057 over its own restricted-shift null, against 0.824 and 0.237 for
cross-correlation and 0.646 and 0.111 for the event-triggered average (for
configurations in levels the excess is conservative; \cref{sec:shift}).
LiNGAM's own null sits at 0.643 and it scores 0.506 within target, so its
AUROC reflects which firms act and which targets are labelled, the
composition the references of \cref{sec:evaluation} are built to expose.
Within target the gap narrows to within the intervals: VAR Granger scores
0.579, against 0.643 for cross-correlation and 0.571 for the event-triggered
average; NTS-NOTEARS falls below chance (0.380), and MiTCD has no excess over
its null. SVAR-RFCI \citep{malinsky2018causal} outputs a PAG
(\cref{app:svar-rfci}): no direct pair receives a $\to$ or $\circ\!\!\to$
mark in any of the 32 cap-2 runs, and almost every adjacency it keeps is bidirected,
which with conditioning sets this small we read as computational stress, not
latent confounding. The marginal test of \cref{sec:protocol} is not in the
table; it scores \aurocObs within action, an excess of \excessShiftFirm, and
\aurocObsWithinTarget within target. It is fixed in advance rather than
selected on the labels. A probe trained on the labels stays within
cross-correlation's intervals (\cref{app:ceiling}).\par}

\begin{table}[t]
    \centering
    \caption{Discovery methods scored as in \cref{sec:evaluation}, with 90\%
    intervals from 2{,}000 bootstrap resamples of actions.
    AUROC: pair-level within-action; excess: over each method's restricted
    per-firm circular-shift null (\cref{sec:shift}; 40 replicates, except
    PCMCI+, 20);
    within target: ranked within target over the labelled pairs (``within
    tgt.''), where an action-blind score gets 0.5. Null replicates refit each
    method's best configuration. Methods are tuned on the full
    label set, so each entry is an upper bound. Bold: best score.
    }
    \label{tab:methods}
    \footnotesize
    \setlength{\tabcolsep}{2pt}
    \begin{tabular*}{\columnwidth}{@{}l@{\extracolsep{\fill}\hspace{6pt}}lccc@{}}
    \toprule
    Method & Family & AUROC & Excess & Within tgt. \\
    \midrule
    PCMCI+ & CB & 0.650 & 0.091 & 0.548 \\
     & & {\scriptsize[0.604, 0.695]} & {\scriptsize[0.045, 0.140]} & {\scriptsize[0.489, 0.606]} \\
    DYNOTEARS & CO & 0.528 & 0.032 & 0.501 \\
     & & {\scriptsize[0.509, 0.547]} & {\scriptsize[0.011, 0.054]} & {\scriptsize[0.456, 0.544]} \\
    NTS-NOTEARS & CO & 0.575 & 0.052 & 0.380 \\
     & & {\scriptsize[0.540, 0.610]} & {\scriptsize[0.025, 0.081]} & {\scriptsize[0.314, 0.447]} \\
    Long.\ LiNGAM & FN & 0.700 & 0.057 & 0.506 \\
     & & {\scriptsize[0.653, 0.747]} & {\scriptsize[0.021, 0.095]} & {\scriptsize[0.458, 0.555]} \\
    VAR Granger & GR & 0.620 & 0.069 & 0.579 \\
     & & {\scriptsize[0.573, 0.667]} & {\scriptsize[0.019, 0.117]} & {\scriptsize[0.528, 0.630]} \\
    MiTCD & GR & 0.541 & 0.000 & 0.482 \\
     & & {\scriptsize[0.514, 0.595]} & {\scriptsize[0.000, 0.001]} & {\scriptsize[0.452, 0.524]} \\
    \midrule
    Cross-correlation & base & \textbf{0.824} & \textbf{0.237} & \textbf{0.643} \\
     & & {\scriptsize[0.780, 0.865]} & {\scriptsize[0.162, 0.309]} & {\scriptsize[0.585, 0.698]} \\
     Event-trig.\ avg.\ & base & 0.646 & 0.111 & 0.571 \\
     & & {\scriptsize[0.588, 0.702]} & {\scriptsize[0.053, 0.168]} & {\scriptsize[0.505, 0.634]} \\
\bottomrule
\end{tabular*}
\end{table}
\subsection{Why do discovery methods struggle?}
\label{sec:failure}

Discovery has to tell direct effects from associations induced by indirect
paths, shared causes and temporal dependence, whether through
conditional-independence tests, prediction or structural assumptions. We ask
whether the distinctions the expert graph implies are recoverable from the
panel.

\textbf{The expert-parent test.}
\label{sec:oracle}
Under the causal Markov and faithfulness conditions, with causal sufficiency
and a correct graph, an action that is not a direct parent of a target is
independent of it given the target's parents, while a direct parent stays
dependent given the other parents \citep{spirtes2000causation}. Conditioning
on a superset of the parents preserves the independence, provided the extra
variables precede the target \citep{runge2019pcmci}. These are implications
of the graph, not of any algorithm, and we test them by supplying the
conditioning sets from the expert graph instead of estimating them. The parent set is conservative: every annotated direct or uncertain
action--state and state--state parent, so no candidate is dropped because its
label was disputed. Extra non-parents cost power, not validity, and uncertain edges
are never scored. For each flow pair and lag $\ell\in\{1,\ldots,4\}$, we
regress the target's level $s_t$ on $a_{t-\ell}$, alone and conditional on
its other candidate parents at lags 1--4, with firm-clustered errors, and
score the pair by $\max_\ell |t_\ell|$ (\cref{app:oracle}). The graph can
still omit parents, place them beyond four quarters or leave common causes
unobserved, so this is a favourable, graph-supplied conditioning test. Raw levels test the graph's implication most directly;
firm scaling, whose denominator uses outcomes across the panel, and first
differences, which change the graph's variables, are complementary checks.

Supplying the expert parents lowers the within-action AUROC in all four
representations (\cref{tab:oracle-auroc}; from 0.647 to 0.533 on raw levels),
and substantial dependence remains among not-direct pairs. After conditioning, 14--44\% of
not-direct pairs remain detected, and 64--83\% of those detected marginally
survive, about as many as direct ones (\cref{tab:oracle-detect}).

A complementary nonlinear test produces the same pattern. We use the
generalised covariance measure with gradient-boosted trees to residualise
both the focal action and target against the expert parents. This relaxes the
linearity assumption for the nuisance relationships, although it tests
residual covariance rather than general conditional independence. The
within-action AUROC still falls from 0.682 to 0.587 on scaled levels and from
0.609 to 0.568 on scaled changes (\cref{tab:oracle-nonlinear}).

As an annotation-sensitivity analysis, we also test whether annotation disagreement has left relevant parents out of the conditioning sets. Since extra preceding variables preserve the graph-implied independence, though they may reduce power, disputed edges can remain candidate parents: a disputed not-direct edge may be a missing true parent, while a disputed direct edge that is truly not-direct is only an
extra control. We therefore score only unanimous pairs and progressively add
the disputed not-direct edges as parents, but discrimination is not
restored (\cref{tab:oracle-disputed}). Demeaning outcomes
by firm and quarter does not restore it either
(\cref{tab:oracle-ablation}). Stock targets remain within 0.05 of chance
within action, with or without the parents (\cref{tab:oracle-stock}).

\begin{table}[t]
\begin{minipage}[c]{\sidewidth{0.53}}
\caption{Expert-parent test on flow targets: AUROC within action (``WA'') and within
target (``WT''), alone and conditioned on the target's other expert parents.
Raw states are as reported; scaled states are divided by the firm's median
total assets. Firm-clustered standard errors; uncertain pairs are in every
conditioning set but not scored (115 direct, 319 not-direct pairs).}
\label{tab:oracle-auroc}
\end{minipage}\sidesep
\begin{minipage}[c]{\sidewidth{0.44}}
\small
\begin{tabular*}{\linewidth}{@{}l@{\extracolsep{\fill}}cccc@{}}
\toprule
& \multicolumn{2}{c}{Marginal} & \multicolumn{2}{c}{Given parents} \\
\cmidrule(lr){2-3}\cmidrule(lr){4-5}
Representation & WA & WT & WA & WT \\
\midrule
Raw level     & 0.647 & 0.622 & 0.533 & 0.523 \\
Scaled level  & 0.677 & 0.576 & 0.595 & 0.549 \\
Raw change    & 0.584 & 0.603 & 0.541 & 0.556 \\
Scaled change & 0.620 & 0.641 & 0.573 & 0.644 \\
\bottomrule
\end{tabular*}
\vspace{-4mm}
\end{minipage}
\end{table}

\textbf{How this relates to the data properties.}
\label{sec:missed}
The test exposes two obstacles tied to \cref{sec:properties}. First, many
direct pairs are invisible even marginally: on firm-scaled states, 30\% are
missed on levels and 59\% on one-quarter changes (\cref{tab:oracle-detect}).
Sparse support explains much of this. In the marginal test, direct-pair
detection rises from \detDirQLow to \detDirQHigh across support quartiles, and
pairs below the support needed to detect a planted effect of 0.1 SD are
detected at \detDirUnderReq, against \detDirOverReq for the rest. Heavy tails
hide more: removing the most extreme 1\% of rows raises direct detection from
\detDirCalled to \detDirTrim.

\label{sec:survive}
Second, not-direct dependence largely survives conditioning on the graph.
On levels, the full parent set shrinks direct coefficients to 0.51 of their
marginal size on raw states and 0.56 on scaled ones, mostly through
same-quarter parents: with bundled actions (\cref{sec:bundling}), co-filed
parents absorb part of the focal action's association. On scaled changes the
parents leave the within-target AUROC unchanged (0.641 to 0.644), so there the
obstacle is power, not bundling. Yet 64--83\% of marginally detected not-direct
associations survive across the four specifications, and 64\% on scaled
levels after demeaning by firm and quarter (\cref{app:oracle}). Nor do they
rest on the filing quarter, whose change can precede the filing: without
it, 71\% of the 73 not-direct pairs detected on scaled changes across all
targets survive (63\% of 148 with it; \cref{app:sensitivity}). Label errors visible in the annotators'
disagreement explain few: Dawid--Skene expects \sensDetNotDs of the
\sensDetNotN to be truly direct (\sensDetNotDsU under a second model). A
confounder explaining 1\% of the residual variance of action and outcome would
erase the median detection, as would one \sensKSizeNot times as strong as log
firm size; the target's own pre-filing change would not at 50 times
(\cref{app:sensitivity}). What remains could be omitted parents (a flow's
own lags are not conditioned on), latent common causes,
or limits of the representation or tests.

\textbf{What this means for the methods.}
\label{sec:predictions}
The expert-parent test bypasses one of the central tasks of discovery:
choosing which variables should be conditioned on. Its candidate parent sets
come directly from the expert graph. The discovery methods receive no such
sets and must recover directness from a panel with sparse interventions,
persistent states, strongly related variables and bundled actions. That the
linear and gradient-boosted tests separate direct from not-direct pairs only
weakly even with graph-supplied parents identifies a difficulty that can
contribute to the limited discrimination of the evaluated method families,
especially within target (\cref{tab:methods}). This is not an impossibility
result: the graph may be incomplete, and other procedures or assumptions
may recover more.

\section{A checklist for a new dataset}
\label{sec:checklist}

The diagnostics of \cref{sec:properties,sec:evaluation} form a checklist for
any panel of rare, dated interventions (\cref{tab:checklist},
\cref{app:checklist}). Before running discovery, measure heavy tails,
persistence, heteroscedasticity, collinearity, the power each intervention's
support allows, and how often interventions are bundled; none of these needs
labels or a method. Before reading a score, compute it under several nulls and
report the excess over the strictest valid one, restrict shifts beyond the
test window, and report a label-only baseline and the units the metric drops.
A label-only baseline takes one line of code, and on
our panel it beats the marginal test and every score in \cref{tab:methods}.

\section{Limitations and conclusion}
\label{sec:limitations}
\label{sec:conclusion}

The panel is balanced by selection, so firms that entered or left are absent.
Actions are extracted by a language model, and an extraction error looks like
an action without effect. The expert graph is partial and may contain errors. Labels are coarse
in time, quarters hide the order of events within them, and binary indicators
ignore the size of an action. The properties and the evaluation use one
linear, additive estimator, so nonlinear or firm-varying effects are outside
what it can see.

On this panel several of the conditions discovery relies on are measurably
violated: extreme tails, \nStatesPersistent of \nStates states close to random
walks, strongly interrelated states and rare interventions. Against references
calibrated for this panel (\cref{sec:evaluation}), the data carry information
about which pairs are direct, yet
standard discovery methods, in the configurations this panel size allows, do
not beat pairwise cross-correlation within action or within target. Most
direct pairs sit on actions too rare to detect a 0.1-SD effect even
marginally, and most
not-direct associations survive the expert parents, so the panel carries
dependence the graph does not account for. The data properties that stress a
method's assumptions can be measured before it is run, with the checklist of
\cref{sec:checklist}.

\bibliography{references}

@book{pearl2009causality,
  author    = {Pearl, Judea},
  title     = {Causality: Models, Reasoning, and Inference},
  edition   = {2nd},
  publisher = {Cambridge University Press},
  year      = {2009}
}

@book{spirtes2000causation,
  author    = {Spirtes, Peter and Glymour, Clark and Scheines, Richard},
  title     = {Causation, Prediction, and Search},
  edition   = {2nd},
  publisher = {MIT Press},
  year      = {2000}
}

@article{runge2019pcmci,
  author  = {Runge, Jakob and Nowack, Peer and Kretschmer, Marlene and Flaxman, Seth and Sejdinovic, Dino},
  title   = {Detecting and quantifying causal associations in large nonlinear time series datasets},
  journal = {Science Advances},
  volume  = {5},
  number  = {11},
  pages   = {eaau4996},
  year    = {2019}
}

@inproceedings{runge2020pcmciplus,
  author    = {Runge, Jakob},
  title     = {Discovering contemporaneous and lagged causal relations in autocorrelated nonlinear time series datasets},
  booktitle = {Proceedings of the 36th Conference on Uncertainty in Artificial Intelligence},
  series    = {PMLR},
  volume    = {124},
  pages     = {1388--1397},
  year      = {2020}
}

@inproceedings{pamfil2020dynotears,
  author    = {Pamfil, Roxana and Sriwattanaworachai, Nisara and Desai, Shaan and Pilgerstorfer, Philip and Georgatzis, Konstantinos and Beaumont, Paul and Aragam, Bryon},
  title     = {{DYNOTEARS}: Structure learning from time-series data},
  booktitle = {Proceedings of the 23rd International Conference on Artificial Intelligence and Statistics},
  series    = {PMLR},
  volume    = {108},
  pages     = {1595--1605},
  year      = {2020}
}

@article{hyvarinen2010varlingam,
  author  = {Hyv{\"a}rinen, Aapo and Zhang, Kun and Shimizu, Shohei and Hoyer, Patrik O.},
  title   = {Estimation of a structural vector autoregression model using non-{G}aussianity},
  journal = {Journal of Machine Learning Research},
  volume  = {11},
  pages   = {1709--1731},
  year    = {2010}
}

@article{granger1969investigating,
  author  = {Granger, Clive W. J.},
  title   = {Investigating causal relations by econometric models and cross-spectral methods},
  journal = {Econometrica},
  volume  = {37},
  number  = {3},
  pages   = {424--438},
  year    = {1969}
}

@article{uhler2013geometry,
  author  = {Uhler, Caroline and Raskutti, Garvesh and B{\"u}hlmann, Peter and Yu, Bin},
  title   = {Geometry of the faithfulness assumption in causal inference},
  journal = {Annals of Statistics},
  volume  = {41},
  number  = {2},
  pages   = {436--463},
  year    = {2013}
}

@inproceedings{faltenbacher2025pc,
  author    = {Faltenbacher, Sofia and Wahl, Jonas and Herman, Rebecca and Runge, Jakob},
  title     = {How {PC}-based methods err: Towards better reporting of assumption violations and small sample errors},
  booktitle = {Proceedings of the 42nd Conference on Uncertainty in Artificial Intelligence},
  series    = {PMLR},
  volume    = {337},
  pages     = {1498--1519},
  year      = {2026}
}

@inproceedings{stein2025causalrivers,
  author    = {Stein, Gideon and Shadaydeh, Maha and Blunk, Jan and Penzel, Niklas and Denzler, Joachim},
  title     = {{CausalRivers}: Scaling up benchmarking of causal discovery for real-world time-series},
  booktitle = {International Conference on Learning Representations},
  year      = {2025}
}

@inproceedings{runge2020c4c,
  author    = {Runge, Jakob and Tibau, Xavier-Andoni and Bruhns, Matthias and Mu{\~n}oz-Mar{\'\i}, Jordi and Camps-Valls, Gustau},
  title     = {The causality for climate competition},
  booktitle = {Proceedings of the NeurIPS 2019 Competition and Demonstration Track},
  series    = {PMLR},
  volume    = {123},
  pages     = {110--120},
  year      = {2020}
}

@inproceedings{mogensen2024industrial,
  author    = {Mogensen, S{\o}ren Wengel and Rathsman, Karin and Nilsson, Per},
  title     = {Causal discovery in a complex industrial system: A time series benchmark},
  booktitle = {Proceedings of the Third Conference on Causal Learning and Reasoning},
  series    = {PMLR},
  volume    = {236},
  pages     = {1218--1236},
  year      = {2024}
}

@inproceedings{reisach2021beware,
  author    = {Reisach, Alexander G. and Seiler, Christof and Weichwald, Sebastian},
  title     = {Beware of the simulated {DAG}! Causal discovery benchmarks may be easy to game},
  booktitle = {Advances in Neural Information Processing Systems},
  volume    = {34},
  year      = {2021}
}

@inproceedings{gentzel2019case,
  author    = {Gentzel, Amanda and Garant, Dan and Jensen, David},
  title     = {The case for evaluating causal models using interventional measures and empirical data},
  booktitle = {Advances in Neural Information Processing Systems},
  volume    = {32},
  year      = {2019}
}

@inproceedings{cheng2024causaltime,
  author    = {Cheng, Yuxiao and Wang, Ziqian and Xiao, Tingxiong and Zhong, Qin and Suo, Jinli and He, Kunlun},
  title     = {{CausalTime}: Realistically generated time-series for benchmarking of causal discovery},
  booktitle = {International Conference on Learning Representations},
  year      = {2024}
}

@inproceedings{gerhardus2020lpcmci,
  author    = {Gerhardus, Andreas and Runge, Jakob},
  title     = {High-recall causal discovery for autocorrelated time series with latent confounders},
  booktitle = {Advances in Neural Information Processing Systems},
  volume    = {33},
  pages     = {12615--12625},
  year      = {2020}
}

@article{huang2020cdnod,
  author  = {Huang, Biwei and Zhang, Kun and Zhang, Jiji and Ramsey, Joseph and Sanchez-Romero, Ruben and Glymour, Clark and Sch{\"o}lkopf, Bernhard},
  title   = {Causal discovery from heterogeneous/nonstationary data},
  journal = {Journal of Machine Learning Research},
  volume  = {21},
  number  = {89},
  pages   = {1--53},
  year    = {2020}
}

@inproceedings{gao2023semistationary,
  author    = {Gao, Shanyun and Addanki, Raghavendra and Yu, Tong and Rossi, Ryan and Kocaoglu, Murat},
  title     = {Causal discovery in semi-stationary time series},
  booktitle = {Advances in Neural Information Processing Systems},
  volume    = {36},
  pages     = {46624--46657},
  year      = {2023}
}

@inproceedings{balsells2025sdci,
  author    = {Balsells-Rodas, Carles and Sumba, Xavier and Narendra, Tanmayee and Tu, Ruibo and Schweikert, Gabriele and Kjellstr{\"o}m, Hedvig and Li, Yingzhen},
  title     = {Causal discovery from conditionally stationary time series},
  booktitle = {Proceedings of the 42nd International Conference on Machine Learning},
  series    = {PMLR},
  volume    = {267},
  pages     = {2715--2741},
  year      = {2025}
}

@article{wiecksosa2025single,
  author  = {Wieck-Sosa, Michael and Haddad, Michel F. C. and Ramdas, Aaditya},
  title   = {The dynamic generalized covariance measure for conditional independence testing with nonstationary time series},
  journal = {arXiv preprint arXiv:2504.21647},
  year    = {2025}
}

@inproceedings{karlsson2023hidden,
  author    = {Karlsson, Rickard K. A. and Krijthe, Jesse H.},
  title     = {Detecting hidden confounding in observational data using multiple environments},
  booktitle = {Advances in Neural Information Processing Systems},
  volume    = {36},
  pages     = {44280--44309},
  year      = {2023}
}

@article{schuemie2014interpreting,
  author  = {Schuemie, Martijn J. and Ryan, Patrick B. and DuMouchel, William and Suchard, Marc A. and Madigan, David},
  title   = {Interpreting observational studies: Why empirical calibration is needed to correct p-values},
  journal = {Statistics in Medicine},
  volume  = {33},
  number  = {2},
  pages   = {209--218},
  year    = {2014}
}

@article{schuemie2016robust,
  author  = {Schuemie, Martijn J. and Hripcsak, George and Ryan, Patrick B. and Madigan, David and Suchard, Marc A.},
  title   = {Robust empirical calibration of p-values using observational data},
  journal = {Statistics in Medicine},
  volume  = {35},
  number  = {22},
  pages   = {3883--3888},
  year    = {2016}
}

@article{schuemie2018empirical,
  author  = {Schuemie, Martijn J. and Hripcsak, George and Ryan, Patrick B. and Madigan, David and Suchard, Marc A.},
  title   = {Empirical confidence interval calibration for population-level effect estimation studies in observational healthcare data},
  journal = {Proceedings of the National Academy of Sciences},
  volume  = {115},
  number  = {11},
  pages   = {2571--2577},
  year    = {2018}
}

@article{ahlmanneltze2025deep,
  author  = {Ahlmann-Eltze, Constantin and Huber, Wolfgang and Anders, Simon},
  title   = {Deep-learning-based gene perturbation effect prediction does not yet outperform simple linear baselines},
  journal = {Nature Methods},
  volume  = {22},
  pages   = {1657--1661},
  year    = {2025}
}

@article{miller2025deep,
  author  = {Miller, Henry E. and Mejia, Gabriel M. and Leblanc, Francis J. A. and Swain, Brendan and Wang, Bo and de Lima Camillo, Lucas Paulo},
  title   = {Deep learning perturbation models can outperform baselines on calibrated metrics},
  journal = {Nature Biotechnology},
  year    = {2026},
  doi     = {10.1038/s41587-026-03307-w}
}

@article{bertrand2004much,
  author  = {Bertrand, Marianne and Duflo, Esther and Mullainathan, Sendhil},
  title   = {How much should we trust differences-in-differences estimates?},
  journal = {Quarterly Journal of Economics},
  volume  = {119},
  number  = {1},
  pages   = {249--275},
  year    = {2004}
}

@article{mackinlay1997event,
  author  = {MacKinlay, A. Craig},
  title   = {Event studies in economics and finance},
  journal = {Journal of Economic Literature},
  volume  = {35},
  number  = {1},
  pages   = {13--39},
  year    = {1997}
}

@article{petersen2009estimating,
  author  = {Petersen, Mitchell A.},
  title   = {Estimating standard errors in finance panel data sets: Comparing approaches},
  journal = {Review of Financial Studies},
  volume  = {22},
  number  = {1},
  pages   = {435--480},
  year    = {2009}
}

@article{cameron2015practitioner,
  author  = {Cameron, A. Colin and Miller, Douglas L.},
  title   = {A practitioner's guide to cluster-robust inference},
  journal = {Journal of Human Resources},
  volume  = {50},
  number  = {2},
  pages   = {317--372},
  year    = {2015}
}

@article{mackinnon2018wild,
  author  = {MacKinnon, James G. and Webb, Matthew D.},
  title   = {The wild bootstrap for few (treated) clusters},
  journal = {The Econometrics Journal},
  volume  = {21},
  number  = {2},
  pages   = {114--135},
  year    = {2018}
}

@article{bell2002bias,
  author  = {Bell, Robert M. and McCaffrey, Daniel F.},
  title   = {Bias reduction in standard errors for linear regression with multi-stage samples},
  journal = {Survey Methodology},
  volume  = {28},
  number  = {2},
  pages   = {169--181},
  year    = {2002}
}

@article{kendall1954note,
  author  = {Kendall, Maurice G.},
  title   = {Note on bias in the estimation of autocorrelation},
  journal = {Biometrika},
  volume  = {41},
  number  = {3/4},
  pages   = {403--404},
  year    = {1954}
}

@article{callaway2021difference,
  author  = {Callaway, Brantly and Sant'Anna, Pedro H. C.},
  title   = {Difference-in-differences with multiple time periods},
  journal = {Journal of Econometrics},
  volume  = {225},
  number  = {2},
  pages   = {200--230},
  year    = {2021}
}

@article{sun2021estimating,
  author  = {Sun, Liyang and Abraham, Sarah},
  title   = {Estimating dynamic treatment effects in event studies with heterogeneous treatment effects},
  journal = {Journal of Econometrics},
  volume  = {225},
  number  = {2},
  pages   = {175--199},
  year    = {2021}
}

@inproceedings{malinsky2018causal,
  author    = {Malinsky, Daniel and Spirtes, Peter},
  title     = {Causal structure learning from multivariate time series in settings with unmeasured confounding},
  booktitle = {Proceedings of the 2018 ACM SIGKDD Workshop on Causal Discovery},
  series    = {Proceedings of Machine Learning Research},
  volume    = {92},
  pages     = {23--47},
  year      = {2018}
}

@inproceedings{sun2023ntsnotears,
  author    = {Sun, Xiangyu and Schulte, Oliver and Liu, Guiliang and Poupart, Pascal},
  title     = {{NTS-NOTEARS}: Learning nonparametric {DBNs} with prior knowledge},
  booktitle = {Proceedings of the 26th International Conference on Artificial Intelligence and Statistics},
  series    = {Proceedings of Machine Learning Research},
  volume    = {206},
  pages     = {1942--1964},
  year      = {2023}
}

@inproceedings{kadowaki2013longitudinal,
  author    = {Kadowaki, Kento and Shimizu, Shohei and Washio, Takashi},
  title     = {Estimation of causal structures in longitudinal data using non-{G}aussianity},
  booktitle = {IEEE International Workshop on Machine Learning for Signal Processing},
  year      = {2013}
}

@article{glymour2019review,
  title={Review of Causal Discovery Methods Based on Graphical Models},
  author={Glymour, Clark and Zhang, Kun and Spirtes, Peter},
  journal={Frontiers in Genetics},
  volume={10},
  pages={524},
  year={2019}
}

@article{assaad2022survey,
  title={Survey and Evaluation of Causal Discovery Methods for Time Series},
  author={Assaad, Charles K and Devijver, Emilie and Gaussier, Eric},
  journal={Journal of Artificial Intelligence Research},
  volume={73},
  pages={767--819},
  year={2022}
}

@article{zheng2018dags,
  title={{DAGs} with {NO TEARS}: Continuous Optimization for Structure Learning},
  author={Zheng, Xun and Aragam, Bryon and Ravikumar, Pradeep K and Xing, Eric P},
  journal={Advances in Neural Information Processing Systems},
  volume={31},
  year={2018}
}

@inproceedings{lachapelle2020gradient,
  title={Gradient-Based Neural {DAG} Learning},
  author={Lachapelle, S{\'e}bastien and Brouillard, Philippe and Deleu, Tristan and Lacoste-Julien, Simon},
  booktitle={International Conference on Learning Representations},
  year={2020}
}

@inproceedings{ferdous2025timegraph,
  title={{TimeGraph}: Synthetic Benchmark Datasets for Robust Time-Series Causal Discovery},
  author={Ferdous, Muhammad Hasan and Hossain, Emam and Gani, Md Osman},
  booktitle={Proceedings of the 31st ACM SIGKDD Conference on Knowledge Discovery and Data Mining},
  year={2025},
  doi={10.1145/3711896.3737439}
}

@inproceedings{herdeanu2025causaldynamics,
  title={{CausalDynamics}: A Large-Scale Benchmark for Structural Discovery of Dynamical Causal Models},
  author={Herdeanu, Benjamin and Nathaniel, Juan and Roesch, Carla and Buch, Jatan and Ramien, Gregor and Haux, Johannes and Gentine, Pierre},
  booktitle={Advances in Neural Information Processing Systems (Datasets and Benchmarks Track)},
  volume={38},
  year={2025}
}

@article{sachs2005causal,
  title={Causal Protein-Signaling Networks Derived from Multiparameter Single-Cell Data},
  author={Sachs, Karen and Perez, Omar and Pe'er, Dana and Lauffenburger, Douglas A and Nolan, Garry P},
  journal={Science},
  volume={308},
  number={5721},
  pages={523--529},
  year={2005}
}

@article{mooij2016tuebingen,
  title={Distinguishing Cause from Effect Using Observational Data: Methods and Benchmarks},
  author={Mooij, Joris M and Peters, Jonas and Janzing, Dominik and Zscheischler, Jakob and Sch{\"o}lkopf, Bernhard},
  journal={Journal of Machine Learning Research},
  volume={17},
  number={32},
  pages={1--102},
  year={2016}
}

@article{chevalley2023causalbench,
  title={A large-scale benchmark for network inference from single-cell perturbation data},
  author={Chevalley, Mathieu and Roohani, Yusuf H. and Mehrjou, Arash and Leskovec, Jure and Schwab, Patrick},
  journal={Communications Biology},
  volume={8},
  pages={412},
  year={2025},
  doi={10.1038/s42003-025-07764-y}
}

@article{zuniga2025fincare,
  title={{FinCARE}: Financial Causal Analysis with Reasoning and Evidence},
  author={Michel, Alejandro and Arun, Abhinav and Sarmah, Bhaskarjit and Pasquali, Stefano},
  journal={arXiv preprint arXiv:2510.20221},
  year={2025}
}

@article{chickering2002optimal,
  title={Optimal Structure Identification with Greedy Search},
  author={Chickering, David Maxwell},
  journal={Journal of Machine Learning Research},
  volume={3},
  pages={507--554},
  year={2002}
}

@article{ban2023causal,
  title={Integrating Large Language Model for Improved Causal Discovery},
  author={Ban, Taiyu and Chen, Lyuzhou and Lyu, Derui and Wang, Xiangyu and Zhu, Qinrui and Tu, Qiang and Chen, Huanhuan},
  journal={IEEE Transactions on Artificial Intelligence},
  volume={6},
  number={11},
  pages={3030--3042},
  year={2025},
  doi={10.1109/TAI.2025.3560927}
}

@article{kiciman2023causal,
  title={Causal Reasoning and Large Language Models: Opening a New Frontier for Causality},
  author={K{\i}c{\i}man, Emre and Ness, Robert and Sharma, Amit and Tan, Chenhao},
  journal={Transactions on Machine Learning Research},
  year={2024}
}

@inproceedings{jiralerspong2024efficient,
  title={Efficient Causal Graph Discovery Using Large Language Models},
  author={Jiralerspong, Thomas and Chen, Xiaoyin and More, Yash and Shah, Vedant and Bengio, Yoshua},
  booktitle={ICLR 2024 Workshop: How Far Are We From AGI},
  year={2024}
}

@inproceedings{roy2025causalllm,
  title={Causal-{LLM}: A Unified One-Shot Framework for Prompt- and Data-Driven Causal Graph Discovery},
  author={Roy, Amartya and Devharish, N. and Ganguly, Shreya and Ghosh, Kripabandhu},
  booktitle={Findings of the Association for Computational Linguistics: EMNLP 2025},
  pages={8259--8279},
  year={2025}
}

@article{chen2025mitcd,
  title={Addressing Information Asymmetry: Deep Temporal Causality Discovery for Mixed Time Series},
  author={Chen, Jiawei and Zhao, Chunhui},
  journal={IEEE Transactions on Pattern Analysis and Machine Intelligence},
  volume={47},
  number={7},
  pages={5723--5741},
  year={2025},
  doi={10.1109/TPAMI.2025.3553957}
}

@book{angrist2009mostly,
  title={Mostly Harmless Econometrics: An Empiricist's Companion},
  author={Angrist, Joshua D and Pischke, J{\"o}rn-Steffen},
  year={2009},
  publisher={Princeton University Press}
}

@article{dawid1979maximum,
  title={Maximum likelihood estimation of observer error-rates using the {EM} algorithm},
  author={Dawid, A. Philip and Skene, Allan M.},
  journal={Journal of the Royal Statistical Society: Series C (Applied Statistics)},
  volume={28},
  number={1},
  pages={20--28},
  year={1979}
}

@article{cinelli2020making,
  title={Making sense of sensitivity: Extending omitted variable bias},
  author={Cinelli, Carlos and Hazlett, Chad},
  journal={Journal of the Royal Statistical Society: Series B (Statistical Methodology)},
  volume={82},
  number={1},
  pages={39--67},
  year={2020}
}

\newpage
\appendix
\setlength{\emergencystretch}{1.5em}
\section{Checklist}
\label{app:checklist}

\Cref{tab:checklist} lists the diagnostics of \cref{sec:checklist} in the
order we would run them. The first six need neither labels nor a discovery
method. Case-study values are reference points, not thresholds, and the
warning signs are qualitative on purpose: a cut-off calibrated on one dataset
would be false precision. Collinearity should be expected in any accounting
panel.

\section{Panel details}
\label{app:data}

The \nActions action types are subtypes of \nItems 8-K items: material
agreements and their termination (1.01, 1.02), completed acquisitions and
disposals (2.01), new direct financial obligations (2.03), exit and disposal
plans (2.05), unregistered equity sales and changes to holders' rights (3.02,
3.03), changes of directors and officers (5.02), amendments to articles or
by-laws (5.03) and other events (8.01). The \nStates states are current and
noncurrent assets, current liabilities, equity, revenues, operating income,
net income, the three cash-flow totals and other comprehensive income.
Because only firms present in every quarter are kept, firms that entered,
exited or were acquired during the window, and the filings that accompany
those events, are not in the panel. The start-of-quarter stamp means $s_t$ is
fixed before any action in quarter $t$ is filed, but a bad quarter can still
prompt an action later in the same quarter.

\begin{table*}[!ht]
\centering
\caption{Diagnostics to run before discovery on a panel. The last three concern
the evaluation.}
\label{tab:checklist}
\scriptsize
\begin{tabular}{@{}>{\raggedright}p{\dimexpr 0.165\linewidth-2\tabcolsep}>{\raggedright}p{\dimexpr 0.215\linewidth-2\tabcolsep}>{\raggedright}p{\dimexpr 0.19\linewidth-2\tabcolsep}>{\raggedright}p{\dimexpr 0.18\linewidth-2\tabcolsep}>{\raggedright\arraybackslash}p{\dimexpr 0.25\linewidth-2\tabcolsep}@{}}
\toprule
Property & Diagnostic & Warning sign & Case study & Response \\
\midrule
Heavy tails & Kurtosis; sum-of-squares share of top 1\% & A few rows carry most of the variation & \topSsWinsMin--\topSsWinsMax after winsorising & Report with and without the top rows; rank or robust tests \\
Persistence & Within-firm AR(1), bias-corrected; ICC & Corrected coefficient near one & \nStatesPersistent states at \rhoHighMin--\rhoHighMax; \rhoKendallAboveOneMin--\rhoKendallAboveOneMax of firms above one after correction & Test changes, not levels; give methods differenced states as well as levels; restrict shift nulls; beware Nickell bias \\
Heteroscedasticity & Auxiliary $R^2$ on scale & A large auxiliary $R^2$, not a rejection & median \rTwoAuxLog & Cluster-robust inference; ignore the rejection rate \\
Collinearity & Computable identities; $R^2$ on the other variables & $R^2$ close to one & up to \colContMax & Drop or combine collinear variables \\
Sparse interventions & Planted-effect power against support & Many interventions below the required support & \nUnderTen of \nUnderTotal below 80\% at 0.1 SD & Report power; pool or drop rare interventions \\
Bundling & Share of firings shared with another intervention & Most firings shared & median \bundleMedian of firings (\cref{sec:bundling}) & Merge or test bundled interventions jointly \\
\midrule
Null choice & Metric under several nulls & Null mean well above 0.5 & \nullLabelPerm--\nullShiftFirmUnr & Report excess over the strictest valid null \\
Shift aliasing & Metric against shift distance & Short shifts score near the observed value & \shiftDistOneMin--\shiftDistOneMax at one quarter & Restrict shifts beyond the test window \\
Label structure & Label-only baseline; units dropped by the metric & Baseline at or above the method & \aurocTargetPrior from target base rates; \nActionsExcluded actions dropped & Report both beside the score \\
\bottomrule
\end{tabular}
\end{table*}

\begin{table}[t]
\centering
\caption{The panel and the expert graph.}
\label{tab:stats}
\footnotesize
\begin{tabular*}{\columnwidth}{@{}l@{\extracolsep{\fill}}r@{}}
\toprule
Firms & \nFirms \\
Quarters (2018Q2--2025Q4) & \nQuarters \\
Firm-quarters & \nFirmQuarters \\
Actions (8-K items) & \nActions (\nItems) \\
States & \nStates \\
State values reported & \stateCompleteness \\
\midrule
Firings per action, median (min--max) & \actFiringsMedian (\actFiringsMin--\actFiringsMax) \\
Firm-quarters with the action, median & \actFiringRateMedian \\
Actions with fewer than 100 firings & \nActionsUnderHundred \\
\midrule
Action--state pairs & \nPairsTotal \\
\quad direct / not direct / uncertain & \nDirect / \nNotDirect / \nUncertain \\
\quad direct: persistent / one-time / no shape & \nDirPersistent / \nDirOneTime / \nDirNoShape \\
Actions with both classes & \nActionsUsable of \nActionsTotal \\
\bottomrule
\end{tabular*}
\end{table}

\begin{table}[h]
\centering
\caption{Action subtypes per 8-K item.}
\label{tab:taxonomy}
\small
\begin{tabular}{@{}llr@{}}
\toprule
Item & Event & Subtypes \\
\midrule
1.01 & Entry into a material definitive agreement & 23 \\
1.02 & Termination of a material definitive agreement & 12 \\
2.01 & Completion of an acquisition or disposition & 4 \\
2.03 & Creation of a direct financial obligation & 5 \\
2.05 & Exit or disposal activities & 4 \\
3.02 & Unregistered sale of equity securities & 1 \\
3.03 & Material modification of holders' rights & 4 \\
5.02 & Departure or appointment of directors and officers & 5 \\
5.03 & Amendment to articles or by-laws; change in fiscal year & 3 \\
8.01 & Other events & 10 \\
\midrule
& Total & \nActions \\
\bottomrule
\end{tabular}
\end{table}

%
\paragraph{Extraction audit.}
We audited the extraction against the filings. Of 30 extracted events,
sampled to over-represent rare subtypes, 24 are supported by the 8-K, 4 are
ambiguous and 2 carry the wrong subtype of the right item; none is invented.
The sample is small: zero of 30 still allows an invention rate of up to about
10\% at 95\% confidence. Against SEC item metadata for 300 firms, about 1.6\%
of firm-quarters miss an event the taxonomy covers, most of the apparent gap
being events that fit no subtype (item 8.01 above all). The embedding-assigned
events were not audited separately, and the audit does not measure how often
one event is split into two subtypes, which would inflate bundling
(\cref{sec:bundling}). A missed or invented event hits an action's direct and
not-direct targets alike, so it pulls the within-action AUROC towards 0.5
rather than inflating it. A wrong subtype is different: it moves an event
to an action with other labels, and its effect on the AUROC has no fixed sign.

\paragraph{State construction.}
Filers report flows cumulatively over the fiscal year, so we take the
discrete-quarter value where the filing reports one and otherwise difference
the year-to-date figures; Q4 is the annual figure minus the nine-month one,
and a revenue difference that comes out negative is left missing. Stocks are
the balance at the quarter end. In an audit of 40 values against the XBRL
filings, all sit in the calendar quarter of their period end, including
fiscal years ending in September, October and January; 31 match the filed
value and 9 could not be verified. None of the 31 is wrong, which bounds the
error rate below about 9\% at 95\% confidence. Some reported values are
exact zeros that are probably unreported rather than true zeros: noncurrent
assets is zero in every quarter for many firms, and other comprehensive
income is often exactly zero. We keep these as reported.

\section{Annotation}
\label{app:labels}

Three annotators labelled each of the \nPairsTotal action--state pairs
independently as a direct cause, not a direct cause, or uncertain; the two
grades of uncertainty they could choose are merged; one annotator left one
pair blank. The annotators disagreed
on \nDisagree pairs.
A financial expert resolved about 250 of them and set out the principles by
which the rest were resolved. Uncertain pairs are held out of every metric. For the pairs judged
direct, the annotators separately judged the shape of the effect, one-time (a
change in a single quarter) or persistent (a change that carries into later
quarters), and the shapes were adjudicated in the same way; \nDirNoShape direct
pairs were not given one. Labels and shapes are judgements about mechanism;
none is estimated from the panel.
%
The annotators followed a written rubric and did not see the panel data. All
three agree on \nUnanimous of the scored pairs: \nUnanimousNot of the \nNotDirect not-direct and
\nUnanimousDir of the \nDirect direct ones, so most direct labels are not unanimous
and the positive class is the noisier one. Fleiss' $\kappa$ over the three
classes is \kappaFleiss on the \nKappaPairs pairs all three labelled.

\section{Nulls}
\label{app:nulls}

Label permutation shuffles labels within action. Firm-block permutation
reassigns whole action histories between firms; the stratified version
reassigns them only within deciles of total action firings, so firm
activity is preserved. Row permutation shuffles
firm-quarters. Circular shifts rotate each action's series within firm, with
one offset per action or one per firm and action, either at any distance or at
least five quarters from the identity. The restricted per-firm shift is
primary (\cref{sec:shift}).

\paragraph{Shift distance.}
Shifting every action by the same distance (\cref{fig:nulls}b) gives
\shiftDistOneMin--\shiftDistOneMax at one quarter, against \aurocObs
observed, \shiftDistTwo at two, averaged over the two directions,
\shiftDistFour at four, \shiftAdmLow--\shiftAdmHigh at five to eight, and
rises only to \shiftDistFifteen at fifteen. An unrestricted shift null draws
its distance at random and so includes the short shifts, which is why it sits
above the restricted one.

\paragraph{Placebo.}
The placebo scores each action against the changes at $h=-4,\dots,-1$, the
changes over quarters $t-5$ to $t-2$ for an action filed in quarter $t$. Its null
is the per-firm shift restricted to at least ten quarters, so that no shift
moves the real effect into the four quarters before the action; it gives
\placeboAuroc against \placeboNull ($p = \placeboP$, \placeboReps
replicates).

\begin{table}[h]
\centering
\caption{Within-action AUROC of the marginal test under each null, from
\nullReps replicates. Excess is the observed \aurocObs minus the null mean. No
replicate of any null reaches the observed value, so every empirical $p$ is at
its floor of \pFloor.}
\label{tab:nulls}
\small
\begin{tabular}{@{}lccc@{}}
\toprule
Null & Mean & SD & Excess \\
\midrule
Firm-block permutation & 0.501 & 0.033 & 0.157 \\
Label permutation & 0.501 & 0.031 & 0.157 \\
Row permutation & 0.514 & 0.032 & 0.144 \\
Stratified firm permutation & 0.527 & 0.034 & 0.131 \\
Shift per action, distance $\ge 5$ & 0.531 & 0.027 & 0.127 \\
Shift per firm, distance $\ge 5$ & 0.538 & 0.027 & 0.120 \\
Shift per action, any distance & 0.549 & 0.028 & 0.109 \\
Shift per firm, any distance & 0.556 & 0.028 & 0.102 \\
\bottomrule
\end{tabular}

\end{table}

\section{Actions with one class}
\label{app:coverage}
The \nZeroDirect actions with no direct target and the \nZeroNotDirect with
no not-direct target drop out of the within-action AUROC. Calling a pair
detected when any horizon is significant at a family-wise 5\%, the median
zero-direct action has \zeroDirectFpr of its pairs flagged
(\zeroDirectFprNull under the primary null), association without directness,
and the median all-direct action has \zeroNotDirectRecall recovered (null
\zeroNotDirectRecallNull).

\section{Planted effects and support}
\label{app:power}

\begin{figure}[h]
\centering
\includegraphics[width=\textwidth]{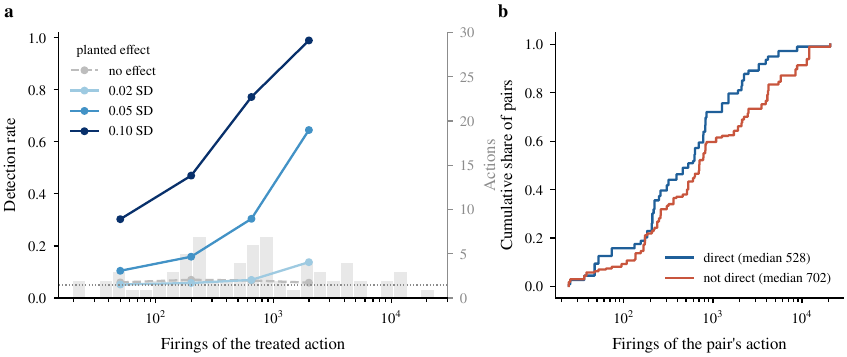}
\caption{(a) Detection of a planted additive effect at 5\% against firm-quarters
treated; grey bars count real actions per support bin (right axis). (b) Support
of the action behind each scored pair, by label.}
\label{fig:power}
\end{figure}

We draw firm-quarters uniformly at random, add $\delta$ standard deviations of
the target's one-quarter change at $h=1$, and run the cell test at 5\%, with
\powReps replicates per point (\cref{fig:power}a). Planted firm-quarters spread
over almost as many firms as firings, while real actions cluster, and under
clustered inference treated firms set power; the grid is therefore
optimistic. \Cref{fig:power}b shows the support behind each scored pair by
label. In the top support quartile \suppTopQNotDirect of pairs are not direct,
against at most \suppOtherQNotDirectMax elsewhere. For \nBundleSameItem of the
\nBundleActions actions the most frequent co-filed partner is a subtype of the
same item, and the least-bundled action still co-occurs in \bundleMin of its
firings.

\section{Heteroscedasticity and accounting relations}
\label{app:extra}
\label{sec:hetero}

A Breusch--Pagan test of the action-to-state residuals against firm size
rejects in all \bpCells cells; with size entered as terciles, the median
statistic is \bpClassicMedian for the
classic test, \bpKoenkerMedian for Koenker's and \bpClusterMedian with
cluster-robust inference. Size explains a median \rTwoAuxTerc of the squared
residuals' variance as terciles and \rTwoAuxLog as log size.

Without totals for assets, liabilities or the change in cash, three standard
identities (total assets as the sum of its parts, the balance sheet, the
cash-flow sum) cannot be computed; the first would be true by construction.
Three relations can be. Current plus noncurrent assets less current
liabilities and equity is implied noncurrent liabilities and should never be
negative; it is in \balNegPct of firm-quarters, which bounds one kind of
extraction error. The change in equity less net income and other
comprehensive income is within 1\% of total assets in \cleanSurplusUnderPct of
firm-quarters, and operating and net income are within 1\% of total assets of
each other in \opNetUnderPct.

\section{Supervised probe}
\label{app:ceiling}
\label{sec:supervised-probe}

Cross-correlation is the strongest score in \cref{tab:methods}. To ask whether
the panel's other label-free statistics add to it, we train a probe that learns
the map from 78 statistics of each pair to the labels, with each action held
out in turn. It is neither a discovery method nor a ceiling on what the panel
holds.

\paragraph{Results.}
The probe reaches 0.841 within action (90\% interval 0.799--0.880), an excess
of 0.143 over its null, against 0.824 for cross-correlation, which is one of
its features and is selected over eight preprocessing choices on the labels
(\cref{sec:setup}). Ranked within target, it reaches 0.681 (0.632--0.733),
above every point estimate in \cref{tab:methods} (0.643 for cross-correlation
and 0.579 for the best discovery method), though cross-correlation lies within
its interval. The other statistics may carry target-specific information that
a single lagged correlation does not.

\paragraph{Features.}
For each action--state pair $(A,S)$, we construct 78 label-free features at
lags $\ell\in\{1,2,3,4\}$: the 39 statistics of
\cref{tab:probe-features}, each on the raw panel and again on the
firm-scaled panel, in which every state is divided by the firm's median total
assets before winsorisation (\cref{sec:setup}). Most use
$\widehat t(Y\sim X)$, the signed homoskedastic Wald statistic for the
coefficient of $X$ in a marginal regression of $Y$ on $X$. We interpret this
statistic as a studentized measure of association rather than as a calibrated
hypothesis test. Applying it to state levels, one-period changes, cumulative
displacements, and firm-demeaned outcomes captures persistent, transitory,
and accumulated responses while separating pooled from within-firm
variation.

We supplement these regression summaries with measures of temporal stability
and lagged alignment. Let $\widehat\beta_{q\ell}$ denote the cross-sectional
slope from regressing $S_{iq}$ on $A_{i,q-\ell}$ across firms in quarter $q$.
Its distribution across quarters captures average strength, dispersion, and
sign coherence. Lagged cross-correlations are computed within each eligible
firm and then averaged across firms, on levels and on one-period changes,
unscaled and firm-scaled; the firm-scaled change variant is the configuration
of \cref{tab:methods}. \Cref{tab:probe-features} gives the full feature set.

\begin{table}[t]
\centering
\small
\begin{tabular}{@{}>{\raggedright}p{0.17\textwidth}cl>{\raggedright\arraybackslash}p{0.33\textwidth}@{}}
\toprule
Feature & Count & Definition & Interpretation \\
\midrule

Level association
& 4
& $\widehat t(S_{it}\sim A_{i,t-\ell})$
& Persistent association visible in the state level. \\

Change association
& 4
& $\widehat t(S_{it}-S_{i,t-1}\sim A_{i,t-\ell})$
& Transitory or flow-like association visible in one-period changes. \\

Cumulative displacement
& 3
& $\widehat t(S_{it}-S_{i,t-\ell}\sim A_{i,t-\ell})$,
  $\ell\in\{2,3,4\}$
& Association with displacement accumulated over multiple quarters.
  Lag one is omitted because it duplicates the one-period change. \\

Firm-demeaned level
& 4
& $\widehat t(\widetilde S_{it}\sim\widetilde A_{i,t-\ell})$
& Within-firm level association after removing time-invariant firm
  differences. \\

Firm-demeaned change
& 4
& $\widehat t(\Delta\widetilde S_{it}\sim
  \widetilde A_{i,t-\ell})$
& Within-firm association with one-period changes. \\

Quarterly-slope mean $t$-statistic
& 4
& $\overline{\widehat\beta}_{\ell}/
  \{\operatorname{sd}(\widehat\beta_{q\ell})/\sqrt{Q_\ell}\}$
& Strength and consistency of the cross-sectional association across eligible
  quarters. \\

Slope dispersion
& 4
& $\operatorname{sd}(\widehat\beta_{q\ell})$
& Temporal heterogeneity in the quarter-specific association. \\

Slope coherence
& 4
& $|\overline{\widehat\beta}_{\ell}|/
  \overline{|\widehat\beta_{q\ell}|}$
& Magnitude-weighted sign consistency across quarters. \\

Lagged cross-correlation
& 4 (8)
& $\displaystyle
  \frac{1}{|\mathcal I_\ell|}
  \sum_{i\in\mathcal I_\ell}
  \left|\operatorname{corr}_{t}
  (A_{i,t-\ell},S_{it})\right|$
& Within-firm temporal alignment, averaged over firms in which both series
  vary. \\

\bottomrule
\end{tabular}
\caption{Label-free observational summaries used by the supervised probe. Here
$\widetilde Z_{it}=Z_{it}-\overline Z_i$,
$Q_\ell$ is the number of eligible quarters, and
$\mathcal I_\ell$ is the set of firms contributing to the lag-$\ell$
cross-correlation. Counts are per panel; the full set applies every row to
the winsorised and the firm-scaled panel, and cross-correlation to levels and
changes on each.}
\label{tab:probe-features}
\end{table}

\paragraph{Training and evaluation.}
We fit a gradient-boosted classifier with 300 trees, learning rate $0.05$,
maximum depth 3, and minimum leaf size 20. The primary evaluation uses
leave-one-action-out cross-fitting. In each fold, all pairs involving one
source action are withheld, the classifier is trained on pairs involving the
remaining actions, and the held-out pairs are scored. Repeating this procedure
over all actions produces one out-of-fold score for each of the 651 strictly
labelled pairs.

We evaluate the probe at the action--state pair level using within-action
and within-target AUROC. Its excess is taken against the same restricted
per-firm shift null as the methods, with 40 replicates, and its interval
treats the null mean as fixed. Each
replicate shifts every action's firings, recomputes the 78 features,
including the four cross-correlation fits, and reruns the full
leave-one-action-out cross-fitting, so the null probe is trained and scored
exactly as the observed one.

\section{Discovery-method protocol}
\label{app:discovery-protocol}

Methods that cannot take missing values run on the panel with the
incomplete state series interpolated linearly within firm; the two firms with
an entirely unobserved series are dropped. A complete-case panel of the firms
with no missing value is the robustness arm. Under firm scaling, states are
additionally divided by the firm's median total assets as in
\cref{sec:panel}; under either scale, they are winsorised at 1\% and
standardised. Firms are never concatenated into one series without a boundary:
lagged rows that would pair one firm's last quarters with the next firm's
first are dropped or masked. PCMCI+ and SVAR-RFCI run with conditioning sets of
at most two variables, a feasibility limit, and binary actions are tested with
a regression-based Wald test rather than partial correlation. Both are run at
four significance levels. PCMCI+ edges are ranked by the strictest level at
which they survive; SVAR-RFCI's PAG marks are reported separately in
\cref{app:svar-rfci}. The event-triggered average scores only the
action-to-state block, and no adapter reads the
labels. Tuned scores are upper bounds.

\subsection{Method configurations}
\label{app:methods}

\paragraph{Shared protocol.}
All methods run on the panel of 2{,}899 firms over 31 quarters.
We search four panel strategies: pooled levels, firm effects (each state
demeaned within firm), firm and quarter effects (two-way demeaning), and
first-differenced states. We cross them with two state scales: raw dollars,
and division by the firm's median total assets as in \cref{sec:panel}. Under
both scales, states are winsorised at 1\% and standardised to zero mean and
unit variance. All transformations apply to the states only; binary action
indicators are neither differenced, demeaned nor rescaled. This gives eight
preprocessing combinations. The lag window is fixed at one to four quarters
for every method. Methods that also estimate contemporaneous edges keep them
in the fit, but only lags 1--4 are scored.

A configuration is scored by pair-level within-action AUROC: each method's
lagged edge scores are collapsed to (action, state) pairs by taking the
strongest lag, uncertain pairs are excluded, and each action's direct pairs
are ranked against its not-direct pairs. Edge-pruning thresholds are never
tuned, since the score ranks every emitted edge. The selected configuration
is the one with the highest score on the full label set, so the reported
numbers are upper bounds over the configurations searched.

\paragraph{Methods with no hyperparameters to tune.}
VAR Granger, longitudinal LiNGAM, cross-correlation and the event-triggered
average have no hyperparameters
beyond the lag window, which is fixed. Each is therefore run once under each
of the eight preprocessing combinations, and the best is reported.
\begin{itemize}
  \item \emph{VAR Granger}: a pooled VAR of every variable on lags 1--4 of
  all variables, fitted within firm so no lag crosses a firm boundary; an
  edge scores $1-p$ for its coefficient.
  \item \emph{Longitudinal LiNGAM}: four lags, default settings.
  \item \emph{Cross-correlation}: the per-firm absolute lagged correlation
  at lags 1--4, averaged over the firms in which both series vary.
  \item \emph{Event-triggered average}: the mean change in the state after
  each firing, with no controls.
\end{itemize}

\paragraph{Methods with tuned hyperparameters.}
DYNOTEARS, NTS-NOTEARS and MiTCD search the panel strategy and state scale
jointly with their own hyperparameters using TPE, with the same budget of 40
configurations each, steered by the same score.
\begin{itemize}
  \item \emph{DYNOTEARS}: $p=4$; $\lambda_W$ and $\lambda_A$ searched on a log
  scale over $[10^{-5},10^{-1}]$. The least-squares term is evaluated through
  the Gram matrix of the lagged design, which minimises the same objective
  as the reference implementation (causalnex) at lower computational cost.
  \item \emph{NTS-NOTEARS}: lag 4, 100 outer iterations, $h_{\text{tol}}=10^{-5}$;
  $\lambda_1$ and $\lambda_2$ searched on a log scale over $[10^{-5},10^{-1}]$.
  Three seeds per configuration, scored by the median.
  \item \emph{MiTCD}: lag 4, 2{,}000 training iterations; learning rate
  $[10^{-4},10^{-1}]$, $\lambda_{\text{sparse}}\in[10^{-3},1]$,
  $\lambda_{\text{ridge}}\in[10^{-4},10^{-1}]$ and
  $\lambda_{\text{rec}}\in[10^{-2},1]$, all on a log scale. Three seeds per
  configuration, scored by the median. Every configuration runs at the full
  iteration budget: a shorter screening stage left low-learning-rate
  configurations unconverged, so its scores could not rank them.
\end{itemize}

\paragraph{Constraint-based methods: computational restrictions.}
PCMCI+ and SVAR-RFCI run a conditional-independence test for every
candidate link and conditioning set, which with 2{,}899 firms and 82
variables is computationally expensive. We therefore make three
restrictions, all for computational reasons.
\begin{enumerate}
  \item \emph{Capped conditioning sets.} PCMCI+ conditions on at most two
  variables in its skeleton phase, and on at most five parents of each of
  the two variables in its MCI phase. SVAR-RFCI conditions on at most two
  variables, tests at most one conditioning set of each size per link, and
  adds no parents of the earlier variable. The upstream defaults are
  unbounded. Small caps limit which confounders the tests can block, so
  these runs test less conditioning than the methods were designed for.
  \item \emph{One fixed configuration.} Beyond the caps, each method is run
  with fixed settings and no hyperparameter search: the regression-based CI
  test (with a Wald variant for binary variables), a lag window of 1--4
  (PCMCI+) or 0--4 (SVAR-RFCI), and, for SVAR-RFCI, its orientation phase
  enabled. Only the preprocessing combination varies.
  \item \emph{Four significance levels.} We use
  $\alpha\in\{0.01,0.1,0.5,0.9\}$ under each preprocessing combination, for
  32 runs per method. For PCMCI+, the four edge sets within a combination are
  pooled into a ranking: an edge scores $1-\alpha^\ast$, where
  $\alpha^\ast$ is the strictest level at which it is retained, and edges
  never retained rank last. This ranking has at most five distinct levels, so
  its AUROC is coarse. SVAR-RFCI instead returns PAG marks, which are retained
  and reported without converting them to a direct-edge score.
\end{enumerate}

\begin{figure}[t]
\centering
\includegraphics[width=0.82\textwidth]{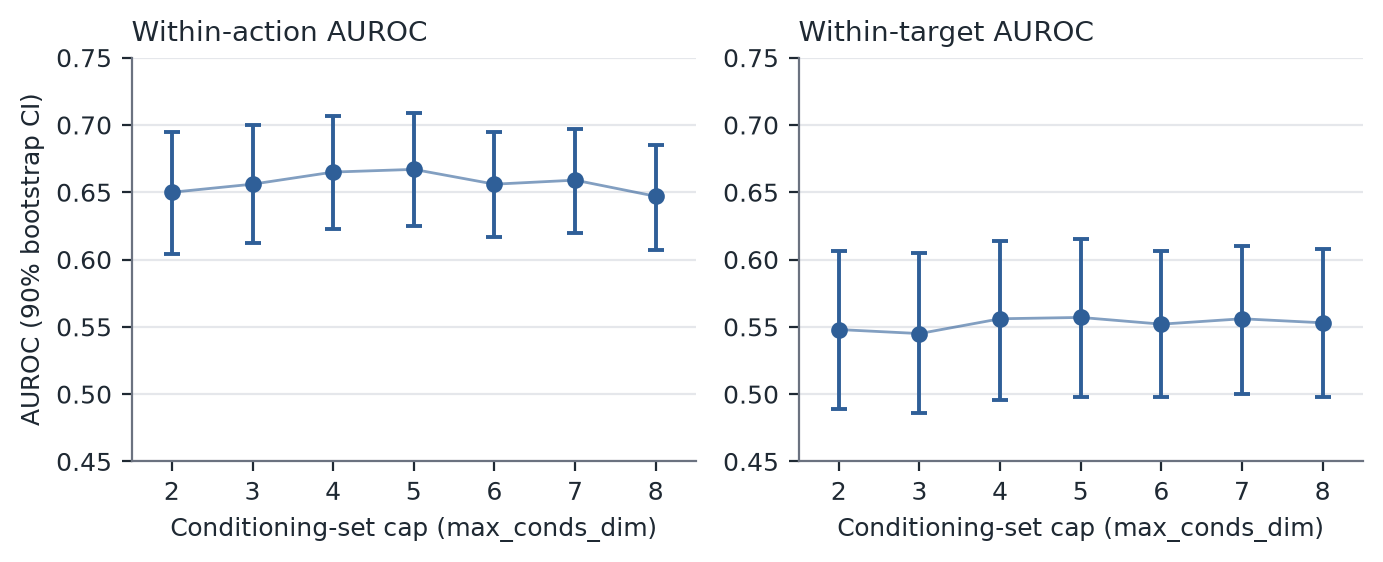}
\caption{PCMCI+ sensitivity to the skeleton conditioning-set cap. The
first-differenced, winsorised variant selected at cap 2 is held fixed while
\texttt{max\_conds\_dim} varies from 2 through 8. Points show within-action
and within-target AUROC; error bars are 90\% bootstrap intervals over
actions. Neither score improves systematically as the cap increases.}
\label{fig:pcmci-cond-cap}
\end{figure}

We test whether the caps drive the results by fixing the preprocessing
variant selected at cap 2 and changing only the relevant conditioning cap.
For PCMCI+, this is the first-differenced, winsorised variant; increasing
\texttt{max\_conds\_dim} from 2 through 8 leaves both evaluation scores nearly
flat, with overlapping 90\% bootstrap intervals
(\cref{fig:pcmci-cond-cap}). Raising this cap alone therefore does not improve
PCMCI+'s discrimination in this sensitivity analysis. The corresponding
SVAR-RFCI sensitivity is reported in \cref{tab:svar-cap-sensitivity}. These
one-factor checks do not evaluate otherwise unrestricted runs: the remaining
method settings stay fixed, and preprocessing is not reselected at each cap.

\begin{table}[t]
\centering
\caption{What each method searches. Preprocessing: four panel strategies
(pooled levels, firm effects, firm and quarter effects, first differences)
crossed with raw or firm-scaled states. Configurations: the number scored.}
\label{tab:method-search}
\footnotesize
\setlength{\tabcolsep}{3pt}
\begin{tabular*}{\columnwidth}{@{}l@{\extracolsep{\fill}}lc@{}}
\toprule
Method & Searched & Configurations \\
\midrule
PCMCI+ & preprocessing only; $\alpha$ swept (4 points) to rank & 8 (32 runs) \\
SVAR-RFCI & preprocessing; $\alpha$ swept; PAG marks reported & 32 runs \\
DYNOTEARS & preprocessing, $\lambda_W$, $\lambda_A$ & 40 (TPE) \\
NTS-NOTEARS & preprocessing, $\lambda_1$, $\lambda_2$ & 40 (TPE) \\
MiTCD & preprocessing, learning rate, three penalties & 40 (TPE) \\
Longitudinal LiNGAM & preprocessing only & 8 \\
VAR Granger & preprocessing only & 8 \\
Cross-correlation & preprocessing only & 8 \\
Event-triggered average & preprocessing only & 8 \\
\bottomrule
\end{tabular*}
\end{table}

\begin{table}[t]
\centering
\caption{Selected configurations behind \cref{tab:methods}. ``Raw'' means no firm scaling. Hyperparameters are shown only for methods for which they were searched.}
\label{tab:selected-method-configs}
\footnotesize
\setlength{\tabcolsep}{3pt}
\begin{tabular*}{\columnwidth}{@{}l@{\extracolsep{\fill}}cll>{\raggedright\arraybackslash}p{0.3\columnwidth}@{}}
\toprule
Method & AUROC & Panel transformation & Scale & Tuned hyperparameters \\
\midrule
PCMCI+ & 0.650 & First differences & Raw & -- \\
DYNOTEARS & 0.528 & Firm and quarter effects & Raw &
$\lambda_W=0.043$, $\lambda_A=0.0026$ \\
NTS-NOTEARS & 0.575 & First differences & Raw &
$\lambda_1=0.0011$, $\lambda_2=0.0013$ \\
Long.\ LiNGAM & 0.700 & First differences & Raw & -- \\
VAR Granger & 0.620 & Firm and quarter effects & Raw & -- \\
MiTCD & 0.541 & Firm effects & Raw &
Learning rate $6.2{\times}10^{-4}$; $\lambda_{\mathrm{sparse}}=0.21$;
$\lambda_{\mathrm{ridge}}=0.0023$; $\lambda_{\mathrm{rec}}=0.14$ \\
Cross-correlation & 0.824 & First differences & Firm-scaled & -- \\
Event-trig.\ avg.\ & 0.646 & Firm and quarter effects & Firm-scaled & -- \\
\bottomrule
\end{tabular*}
\end{table}

\subsection{SVAR-RFCI PAG output}
\label{app:svar-rfci}

SVAR-RFCI is evaluated on the marks it outputs rather than mapped into the
direct-edge score of \cref{tab:methods}. Its output is almost entirely
bidirected. In every preprocessing combination and significance
level, every lagged adjacency it keeps is marked $\leftrightarrow$, apart
from at most eight $\to$ or $\circ\!\!\to$ marks in the whole PAG
(\cref{tab:svar-rfci-marks}). Loosening $\alpha$ only adds more: at $\alpha=0.9$, about 17{,}000--18{,}600 of the 26{,}896
possible lagged cells are bidirected, as are almost all labelled pairs.

\begin{table}[t]
\centering
\caption{SVAR-RFCI output by preprocessing, panel transformation and
significance level. Left: lagged edge marks in the whole PAG (82 variables,
lags 1--4; 26{,}896 possible cells). Right: labelled action--state pairs with a
bidirected ($\leftrightarrow$) mark at any lag 1--4, or not adjacent at any
lag. No labelled pair receives a $\to$ or $\circ\!\!\to$ mark except one
not-direct pair at $\alpha=0.01$ under firm effects and under firm and quarter
effects (winsorised). ``Winsorised'': raw-dollar states trimmed 1\% over all
firm-quarters; ``firm-scaled'': states divided by the firm's median total
assets and trimmed 1\% within each quarter. Both are standardised.}
\label{tab:svar-rfci-marks}
\scriptsize
\setlength{\tabcolsep}{2.5pt}
\begin{tabular*}{\columnwidth}{@{}ll@{\extracolsep{\fill}}rrrcrrcrr@{}}
\toprule
 & & \multicolumn{3}{c}{Whole PAG} & & \multicolumn{2}{c}{Direct (222)} & & \multicolumn{2}{c}{Not-direct (429)} \\
\cmidrule{3-5}\cmidrule{7-8}\cmidrule{10-11}
Transformation & $\alpha$ & $\leftrightarrow$ & $\to$ & $\circ\!\!\to$ & & $\leftrightarrow$ & none & & $\leftrightarrow$ & none \\
\midrule
\multicolumn{11}{@{}l}{\emph{Winsorised}} \\
Pooled & 0.01 & 3{,}653 & 2 & 0 & & 62 & 160 & & 107 & 322 \\
       & 0.1  & 6{,}011 & 2 & 1 & & 94 & 128 & & 186 & 243 \\
       & 0.5  & 11{,}122 & 0 & 0 & & 172 & 50 & & 347 & 82 \\
       & 0.9  & 18{,}576 & 0 & 0 & & 221 & 1 & & 426 & 3 \\
Firm effects & 0.01 & 2{,}651 & 2 & 2 & & 48 & 174 & & 76 & 352 \\
             & 0.1  & 4{,}743 & 0 & 0 & & 81 & 141 & & 153 & 276 \\
             & 0.5  & 9{,}765 & 0 & 0 & & 166 & 56 & & 333 & 96 \\
             & 0.9  & 17{,}673 & 0 & 0 & & 219 & 3 & & 429 & 0 \\
Firm and quarter & 0.01 & 2{,}600 & 2 & 2 & & 53 & 169 & & 72 & 356 \\
effects          & 0.1  & 4{,}681 & 0 & 0 & & 84 & 138 & & 151 & 278 \\
                 & 0.5  & 9{,}703 & 0 & 0 & & 156 & 66 & & 332 & 97 \\
                 & 0.9  & 17{,}648 & 0 & 0 & & 219 & 3 & & 427 & 2 \\
First-differenced & 0.01 & 2{,}416 & 3 & 5 & & 64 & 158 & & 74 & 355 \\
                  & 0.1  & 4{,}431 & 0 & 0 & & 98 & 124 & & 157 & 272 \\
                  & 0.5  & 9{,}398 & 0 & 0 & & 166 & 56 & & 324 & 105 \\
                  & 0.9  & 17{,}306 & 0 & 0 & & 220 & 2 & & 425 & 4 \\
\midrule
\multicolumn{11}{@{}l}{\emph{Firm-scaled}} \\
Pooled & 0.01 & 3{,}566 & 3 & 2 & & 62 & 160 & & 129 & 300 \\
       & 0.1  & 5{,}833 & 0 & 0 & & 100 & 122 & & 225 & 204 \\
       & 0.5  & 11{,}072 & 0 & 0 & & 169 & 53 & & 383 & 46 \\
       & 0.9  & 18{,}493 & 0 & 0 & & 219 & 3 & & 427 & 2 \\
Firm effects & 0.01 & 2{,}826 & 4 & 1 & & 58 & 164 & & 107 & 322 \\
             & 0.1  & 5{,}063 & 0 & 0 & & 93 & 129 & & 208 & 221 \\
             & 0.5  & 10{,}228 & 0 & 0 & & 160 & 62 & & 366 & 63 \\
             & 0.9  & 17{,}936 & 0 & 0 & & 220 & 2 & & 427 & 2 \\
Firm and quarter & 0.01 & 2{,}787 & 4 & 1 & & 61 & 161 & & 107 & 322 \\
effects          & 0.1  & 5{,}008 & 0 & 0 & & 95 & 127 & & 208 & 221 \\
                 & 0.5  & 10{,}179 & 0 & 0 & & 170 & 52 & & 364 & 65 \\
                 & 0.9  & 17{,}957 & 0 & 0 & & 217 & 5 & & 428 & 1 \\
First-differenced & 0.01 & 2{,}499 & 4 & 4 & & 70 & 152 & & 118 & 311 \\
                  & 0.1  & 4{,}624 & 0 & 0 & & 112 & 110 & & 208 & 221 \\
                  & 0.5  & 9{,}767 & 0 & 0 & & 184 & 38 & & 374 & 55 \\
                  & 0.9  & 17{,}471 & 0 & 0 & & 219 & 3 & & 427 & 2 \\
\bottomrule
\end{tabular*}
\end{table}

A bidirected mark $A\leftrightarrow Y$ asserts that neither variable is an
ancestor of the other, so their dependence is attributed to a latent common
cause. An expert-direct pair, by definition, is one in which the action is an
ancestor of the state. In the reported cap-2 runs, SVAR-RFCI never recovers
this ancestry: no direct pair receives a $\to$ or $\circ\!\!\to$ mark.
Whenever it keeps a direct pair adjacent, it instead marks the pair
bidirected, contradicting the label.

The bidirected marks also fail to distinguish direct from not-direct pairs.
At $\alpha=0.5$ under winsorised pooled levels, they cover 77\% of direct
pairs and 81\% of not-direct pairs, and the two shares remain close across
settings.
SVAR-RFCI therefore fails to recover direct edges not by abstaining, but by
marking the action--state dependencies it keeps as bidirected.

\paragraph{Conditioning-cap sensitivity.}
We hold fixed the pooled, winsorised variant selected at cap 2 and vary only
\texttt{max\_p\_global} from 2 to 5
(\cref{tab:svar-cap-sensitivity}). Higher caps prune more adjacencies, but do
not recover orientations for expert-direct pairs. Across all cap and
significance-level combinations, every direct pair is either marked
bidirected or absent, except for one pair that gains a $\circ\!\!\to$ mark at
caps 4 and 5 under $\alpha=0.01$. At $\alpha=0.9$, 221 of the 222 direct pairs
are bidirected and one is absent at every cap. A directed-edge adapter
therefore remains at 0.5 AUROC at every cap.

\begin{table}[t]
\centering
\caption{SVAR-RFCI sensitivity to the global conditioning cap for the 222
expert-direct pairs. The pooled, winsorised variant selected at cap 2 is held
fixed while \texttt{max\_p\_global} varies. Each cell reports the number
marked bidirected / not adjacent / oriented at any lag 1--4; ``oriented''
combines $\to$ and $\circ\!\!\to$.}
\label{tab:svar-cap-sensitivity}
\small
\begin{tabular*}{\columnwidth}{@{}c@{\extracolsep{\fill}}cccc@{}}
\toprule
Cap & $\alpha=0.01$ & $\alpha=0.1$ & $\alpha=0.5$ & $\alpha=0.9$ \\
\midrule
2 & 62 / 160 / 0 & 94 / 128 / 0 & 172 / 50 / 0 & 221 / 1 / 0 \\
3 & 57 / 165 / 0 & 89 / 133 / 0 & 171 / 51 / 0 & 221 / 1 / 0 \\
4 & 53 / 168 / 1 & 85 / 137 / 0 & 164 / 58 / 0 & 221 / 1 / 0 \\
5 & 49 / 172 / 1 & 79 / 143 / 0 & 160 / 62 / 0 & 221 / 1 / 0 \\
\bottomrule
\end{tabular*}
\end{table}

The cap-2 choice alone therefore does not explain SVAR-RFCI's result:
increasing it does not recover directed separation. The sensitivity still
holds the other computational restrictions fixed, including one conditioning
set per size.

\section{Expert-parent test}
\label{app:oracle}

The expert-parent test of \cref{sec:oracle} uses the four representations in
\cref{tab:oracle-auroc}. Scaled states are divided by the firm's median total
assets and winsorised at 1\% within each quarter, as in \cref{sec:panel}; raw
states are reported in dollars and winsorised at 1\% over all firm-quarters.
Change scores difference the winsorised states. Each specification uses one
complete-case sample for all of its fits: 63{,}288
firm-quarters for levels and 60{,}233 for changes. Actions are binary
throughout. The conditioning set holds the target's other expert parents at
every lag 1--4, uncertain parents included. Only the focal action at the lag
being tested is excluded; its other lags stay in the set when it is a parent.
For a stock, the set also contains its own lagged level. Uncertain pairs are
not scored. Adding earlier variables to a set that already contains the
parents cannot break the separation of a not-direct pair, so including
uncertain parents is conservative. The detection cut-off, 2.49, is the
\v{S}id\'ak level for a family-wise 5\% over four lags. Standard errors are
clustered by firm, with $t$ on $G-1$ degrees of freedom.
\Cref{tab:oracle-detect} gives detection rates by class for flow targets, and
\cref{tab:oracle-timing} conditions on one timing class of parents at a time.

\paragraph{Nonlinear conditional association.}
We repeat the test with the generalised covariance measure. The focal action
lag and target are each regressed on the conditioning set using
gradient-boosted trees (depth 4, 100 trees) with three firm-grouped
cross-fitting folds. The score is the standardised mean product of their
out-of-fold residuals, maximised over lags 1--4. The rows, pairs and
conditioning sets are the same as in the linear test; only the
residualisation changes. This tests residual covariance after flexible
adjustment, not general conditional independence. It yields the same ordering
(\cref{tab:oracle-nonlinear}).

\begin{table}[h]
\centering
\caption{Generalised covariance measure on flow targets with nonlinear
nuisance regressions: AUROC within action / within target, alone and
conditioned on the expert parents. Gradient-boosted residualisation with three
firm-grouped cross-fitting folds; uncertain pairs are in every conditioning
set but not scored.}
\label{tab:oracle-nonlinear}
\small
\begin{tabular*}{\columnwidth}{@{}l@{\extracolsep{\fill}}cccc@{}}
\toprule
Conditioning set & Raw level & Raw change & Scaled level & Scaled change \\
\midrule
None (marginal) & 0.641 / 0.602 & 0.611 / 0.599 & 0.682 / 0.605 & 0.609 / 0.647 \\
Expert parents  & 0.505 / 0.469 & 0.562 / 0.588 & 0.587 / 0.533 & 0.568 / 0.622 \\
\bottomrule
\end{tabular*}
\end{table}

\begin{table}[h]
\centering
\caption{Expert-parent test on flow targets: share of pairs detected alone and given
the target's other expert parents, and the share of pairs detected alone that
stay detected. A pair is detected if $\max_\ell|t_\ell|>2.49$, a family-wise
5\% over four lags, with firm-clustered standard errors. Direct: 115 pairs;
not direct: 319.}
\label{tab:oracle-detect}
\small
\begin{tabular*}{\columnwidth}{@{}l@{\extracolsep{\fill}}cccccc@{}}
\toprule
& \multicolumn{3}{c}{Direct} & \multicolumn{3}{c}{Not direct} \\
\cmidrule(lr){2-4}\cmidrule(lr){5-7}
Specification & Alone & Given parents & Kept & Alone & Given parents & Kept \\
\midrule
Scaled level  & 70.4\% & 53.9\% & 70\% & 50.5\% & 43.9\% & 83\% \\
Scaled change & 40.9\% & 33.0\% & 74\% & 20.7\% & 16.0\% & 64\% \\
Raw level     & 73.9\% & 56.5\% & 76\% & 50.5\% & 41.7\% & 78\% \\
Raw change    & 26.1\% & 25.2\% & 83\% & 16.9\% & 13.8\% & 70\% \\
\bottomrule
\end{tabular*}
\end{table}

\begin{table}[h]
\centering
\caption{Flow targets: conditioning on one timing class of the expert parent
set at a time. Relative to the focal action at lag $\ell$, pre-source,
same-time and post-source parents occur at lags $>\ell$, $=\ell$ and
$<\ell$. Each cell gives the AUROC within action / within target and, in
brackets, the median ratio of the direct pairs' $|\hat\beta|$ to the marginal
fit. Uncertain pairs are in every conditioning set but not scored.}
\label{tab:oracle-timing}
\small
\begin{tabular*}{\columnwidth}{@{}l@{\extracolsep{\fill}}cccc@{}}
\toprule
Parents & Scaled level & Scaled change & Raw level & Raw change \\
\midrule
Pre-source  & 0.679 / 0.595 (0.87) & 0.611 / 0.626 (1.00) & 0.633 / 0.610 (0.84) & 0.571 / 0.595 (1.00) \\
Same-time   & 0.636 / 0.548 (0.76) & 0.592 / 0.644 (1.03) & 0.554 / 0.552 (0.70) & 0.530 / 0.558 (1.01) \\
Post-source & 0.660 / 0.579 (0.87) & 0.622 / 0.639 (1.00) & 0.633 / 0.611 (0.82) & 0.560 / 0.579 (1.00) \\
\bottomrule
\end{tabular*}
\end{table}

Under the full parent set, the median direct coefficient retains 0.51 of its
marginal magnitude on raw levels and 0.56 on scaled levels, against 1.04 and
1.03 on raw and scaled changes. The corresponding median standard-error
ratios are 0.91 on raw levels and 1.09 and 1.07 on raw and scaled changes.
For not-direct pairs, coefficient ratios are 0.91 on raw levels and
0.97--0.99 on changes; standard-error ratios on changes are 1.00--1.01.

\paragraph{Parents the graph may be missing.}
Disagreement among annotators marks edges the graph may have wrong. We fix
the evaluation set to the 343 unanimously labelled flow pairs (59 direct, 191
within-action comparisons) and add to the parent sets, at lags 1--4, the 35
not-direct flow edges with at least one direct or uncertain vote, in 100
random nested orders from none to all of them. Adding them never restores
discrimination (\cref{tab:oracle-disputed}): the level loss shrinks, the
change loss stays at 0.05--0.06, direct coefficients barely move, and
64--79\% of the not-direct detections still survive with all 35 added. The
ranges over the 100 orders are within 0.05 of AUROC and 6 points of retention
at every step.

\begin{table}[h]
\centering
\caption{Adding the 35 disputed not-direct flow edges to the parent sets,
none $\to$ all, on the 343 unanimously labelled flow pairs. Loss: fall in
within-action AUROC from conditioning. Kept: share of not-direct detections
that survive conditioning. Ratio: median direct $|\hat\beta|$ given the
parents over its marginal value.}
\label{tab:oracle-disputed}
\small
\begin{tabular*}{\columnwidth}{@{}l@{\extracolsep{\fill}}ccc@{}}
\toprule
Specification & Loss & Kept & Ratio \\
\midrule
Scaled level  & 0.094 $\to$ 0.068 & 83\% $\to$ 79\% & 0.59 $\to$ 0.55 \\
Scaled change & 0.047 $\to$ 0.058 & 64\% $\to$ 64\% & 1.05 $\to$ 1.07 \\
Raw level     & 0.126 $\to$ 0.094 & 81\% $\to$ 78\% & 0.56 $\to$ 0.53 \\
Raw change    & 0.052 $\to$ 0.058 & 75\% $\to$ 75\% & 1.05 $\to$ 1.05 \\
\bottomrule
\end{tabular*}
\end{table}

\paragraph{Firm and quarter demeaning.}
We rerun the test with the states two-way demeaned before the fit and the
actions left as filed (\cref{tab:oracle-ablation}). Conditioning on the expert
parents lowers the within-action AUROC on raw levels, while slightly raising
it on raw changes; on both scaled representations, which the table does not
show, it also lowers it. It again fails to separate the classes. On scaled
levels, demeaning the state outcomes lowers
detection in both classes, but 64\% of the not-direct detections (32\% to
24\%) and 75\% of the direct detections (45\% to 37\%) survive the parents.

\begin{table}[h]
\centering
\caption{Expert-parent test on flow targets with raw state outcomes demeaned by firm
and quarter: AUROC within action (``WA'') and within target (``WT''), alone
and conditioned on the target's other expert parents. The states are exactly
two-way demeaned after choosing the level/change representation; actions
stay as filed. Firm-clustered standard errors; uncertain pairs are in every
conditioning set but not scored.}
\label{tab:oracle-ablation}
\small
\begin{tabular}{@{}lcccc@{}}
\toprule
& \multicolumn{2}{c}{Level} & \multicolumn{2}{c}{Change} \\
\cmidrule(lr){2-3}\cmidrule(lr){4-5}
Conditioning set & WA & WT & WA & WT \\
\midrule
None (marginal) & 0.647 & 0.649 & 0.549 & 0.601 \\
Expert parents  & 0.538 & 0.579 & 0.557 & 0.556 \\
\bottomrule
\end{tabular}
\end{table}

\paragraph{Stock targets.}
Stock targets do not separate the classes even marginally under either score,
and conditioning on the expert parents leaves them close to 0.5
(\cref{tab:oracle-stock}). Their conditioning sets also hold the stock's own
lagged level.

\begin{table}[h]
\centering
\caption{Expert-parent test on stock targets with raw states: AUROC within action
(``WA'') and within target (``WT''), alone and conditioned on the target's
other expert parents. Firm-clustered standard errors; uncertain pairs are in
every conditioning set but not scored (217 pairs).}
\label{tab:oracle-stock}
\small
\begin{tabular}{@{}lcccc@{}}
\toprule
& \multicolumn{2}{c}{Level} & \multicolumn{2}{c}{Change} \\
\cmidrule(lr){2-3}\cmidrule(lr){4-5}
Conditioning set & WA & WT & WA & WT \\
\midrule
None (marginal) & 0.452 & 0.395 & 0.516 & 0.597 \\
Expert parents  & 0.532 & 0.532 & 0.516 & 0.516 \\
\bottomrule
\end{tabular}
\end{table}

\section{Size of the cell and pair tests}
\label{app:inference}

\Cref{tab:size} reports the rejection rate of each inference convention under a
true null, from \sizeReps simulated cells per row, at a nominal 5\%, so each
rate carries a Monte Carlo standard error of about 0.01. Each
simulated panel has 400 firms over 20 quarters, and each treated firm fires
the action four times. The outcome is Gaussian, a firm effect plus independent noise, so the table
measures size on well-behaved data; the panel's heavy tails and exact zeros
can change it. The treated count is the number of firms that fire the
simulated action. The cell
$t$ test with $\min(G_1, G_0) - 1$ degrees of freedom, which we use, rejects at
between \sizeRuleMin and \sizeRuleMax across the grid. Referencing the same
statistic to $G-1$ degrees of freedom rejects at \sizeNaiveTen with ten treated
firms. The Bell--McCaffrey correction \citep{bell2002bias} and the restricted subcluster wild bootstrap, with weights
drawn per firm-quarter rather than per firm \citep{mackinnon2018wild}, are about as well sized as our rule. The joint Wald
test over four horizons is not: it rejects at \sizeWaldTen with ten treated
firms and falls to \sizeWaldEighty only at 80. We therefore flag the
\nWaldUnreliable pairs whose action fires in fewer than 80 firms; they remain
in the ranking.

%
\paragraph{Ordinary standard errors.}
On the panel itself, refitting the marginal test with ordinary instead of
firm-clustered standard errors barely moves the median $|t|$ but changes
significance in \naiveFlipShare of cells. The overall rejection rate is unchanged, but
detections move from direct to not-direct pairs (\detDirClustered to \detDirNaive and \detNotClustered to
\detNotNaive), and the AUROC of each pair's smallest $p$ value falls from \aurocMinPClustered to
\aurocMinPNaive. Clustering corrects within-firm dependence and unequal variance
together, so the two are not separated here.

\begin{table}[h]
\centering
\caption{Rejection rates under the null at a nominal 5\%, on simulated
Gaussian outcomes.}
\label{tab:size}
\small
\begin{tabular}{@{}rccccc@{}}
\toprule
Treated & $t$, $G_1{-}1$ df & $t$, $G{-}1$ df & CR2-BM & Wild boot. & Joint $F$ \\
\midrule
10 & 0.050 & 0.074 & 0.048 & 0.044 & 0.156 \\
20 & 0.052 & 0.074 & 0.046 & 0.048 & 0.084 \\
30 & 0.048 & 0.056 & 0.046 & 0.042 & 0.070 \\
40 & 0.034 & 0.050 & 0.032 & 0.032 & 0.098 \\
80 & 0.056 & 0.060 & 0.056 & 0.050 & 0.062 \\
\bottomrule
\end{tabular}

\end{table}

\section{Choice of pair score}
\label{app:aggregation}

We considered four pair scores: the smallest per-horizon $p$ value, the
largest and the mean per-horizon $|t|$, and the joint Wald test. The rule fixed
in advance was to take the score with the largest excess over its own
restricted per-firm shift null. The rule was applied to the pairs on which the annotators agreed;
\Cref{tab:aggregation} gives the same comparison on all scored pairs. The joint Wald test also has the largest
excess there, and the four within-action AUROCs span \aurocAggMin to
\aurocAggMax, so no conclusion in the paper depends on the choice.

\begin{table}[h]
\centering
\caption{Pair scores on all scored pairs. The shift null is the restricted
per-firm shift. $^\ast$Score used throughout.}
\label{tab:aggregation}
\small
\begin{tabular}{@{}lcccc@{}}
\toprule
Pair score & Within-action & Pooled & Shift null & Excess \\
\midrule
Minimum $p$ & 0.640 & 0.601 & 0.533 & 0.106 \\
Maximum $|t|$ & 0.638 & 0.604 & 0.533 & 0.104 \\
Mean $|t|$ & 0.644 & 0.613 & 0.537 & 0.107 \\
Joint Wald$^\ast$ & 0.658 & 0.609 & 0.538 & 0.120 \\
\bottomrule
\end{tabular}

\end{table}

\section{Sensitivity to the labels, the inference and confounding}
\label{app:sensitivity}

The labels are expert judgements, so the headline measures how far the panel
agrees with them. Here we ask how much that agreement depends on which
judgements are trusted, on the inference, and on unobserved confounding. The
score is the marginal test of \cref{sec:protocol}, ranked within action unless
stated. The reference for a relabelled set is the 95th percentile of the
restricted per-firm shift null under the final labels, \sensNullQ; the panel
nulls are not recomputed for each relabelling, except for the unanimous
subset below, but the label-permutation null in \cref{tab:label-sens} is.

\paragraph{Which labels are trusted.}
On the \sensUnanPairs unanimously labelled pairs the AUROC is \sensUnanAuroc,
an excess of \sensUnanExcess over the restricted shift null recomputed on that
subset ($p=\sensUnanP$).
The pair score was chosen on this subset (\cref{app:aggregation}), but on all
scored pairs the four candidates span only \aurocAggMin--\aurocAggMax. Dropping the \sensContestedPairs pairs with
a vote for the other class raises the AUROC to \sensNoOppAuroc; those pairs
alone form only \sensContestedComparisons comparisons, too few to score. Each
annotator's labels alone give \sensAnnotMin--\sensAnnotMax, and the three ways
of scoring the uncertain pairs give \sensUncDirAuroc--\sensUncVoteAuroc
(\cref{tab:label-sens}). Only the strictest definition, which turns the
\sensStrictRelabelled non-unanimous direct pairs into not-direct ones, falls
to its label-permutation reference (\sensStrictAuroc). Ranked within target,
the unanimous subset scores \sensUnanWT; with only \nUnanimousDir direct
pairs, it is too small to resolve the signal shared
across a target's actions (\cref{sec:identity}).

\begin{table}[h]
\centering
\caption{The marginal test under alternative label sets. Within-action AUROC,
the 95th percentile of a label-permutation null recomputed for each set, and
the within-target AUROC. Uncertain pairs are left out except in the last
block.}
\label{tab:label-sens}
\footnotesize
\begin{tabular}{@{}lrrrccc@{}}
\toprule
Labels & Pairs & Direct & Comparisons & Within-action & Label perm.\ q95 & Within tgt. \\
\midrule
Final labels (headline) & 651 & 222 & 869 & 0.658 & 0.547 & 0.564 \\
\midrule
Unanimous pairs only & 442 & 95 & 340 & 0.647 & 0.589 & 0.516 \\
No vote for the other class & 594 & 185 & 652 & 0.687 & 0.555 & 0.592 \\
\midrule
Direct if $\ge 2$ direct votes & 651 & 200 & 861 & 0.653 & 0.545 & 0.559 \\
Direct if $\ge 1$ direct vote & 651 & 240 & 883 & 0.675 & 0.546 & 0.581 \\
Direct only if all 3 vote direct & 651 & 95 & 601 & 0.566 & 0.564 & 0.541 \\
Annotator A1 alone & 557 & 138 & 599 & 0.668 & 0.556 & 0.572 \\
Annotator A2 alone & 587 & 181 & 738 & 0.649 & 0.564 & 0.563 \\
Annotator A3 alone & 613 & 216 & 805 & 0.655 & 0.559 & 0.559 \\
\midrule
Uncertain counted direct & 781 & 352 & 1,358 & 0.643 & 0.555 & 0.563 \\
Uncertain counted not direct & 781 & 222 & 1,184 & 0.644 & 0.547 & 0.589 \\
Uncertain by vote majority & 781 & 267 & 1,280 & 0.658 & 0.539 & 0.588 \\
\bottomrule
\end{tabular}

\end{table}

\paragraph{How many labels would have to be wrong.}
Under random flips, the mean AUROC falls to the original benchmark's
95th-percentile reference, \sensNullQ, only once \sensFlipNotToDir of
not-direct labels are made direct, or \sensFlipBoth of each class are
swapped; flipping direct labels to not direct never gets there up to
\sensFlipMaxRate. The null can itself move when the class configuration
changes, so these are breakdown points against the benchmark's reference,
not against a null recomputed for each hypothetical graph. An adversary that flips the most damaging label at each
step needs \sensAdvFlips flips (\sensAdvShare of scored pairs, all not direct
to direct), or \sensAdvFlipsNonUnan if it may flip only non-unanimous pairs.
That is a worst case over arbitrary errors. A Dawid--Skene model of the votes
\citep{dawid1979maximum} reproduces the final label for \sensDsAgree of pairs
(\sensDsUAgree with uncertain votes as a third response). Restricted to the
\sensAdvDsPool pairs it considers plausibly mislabelled (\sensAdvDsUPool under
the second model), the adversary stops at \sensAdvDsAuroc (\sensAdvDsUAuroc),
and labels redrawn from its posteriors give a mean of \sensDsMean, with
2.5\% quantile \sensDsLow (\sensDsUMean and \sensDsULow).

\paragraph{Inference and scaling.}
Clustering by firm and by quarter, which also absorbs shocks shared across
firms within a quarter, leaves the AUROC of the smallest per-horizon $p$
(\cref{app:inference}) at \sensTwFqAuroc against \sensTwFirmAuroc, and moves
the within-target AUROC from \sensTwFirmWT to \sensTwFqWT. The median standard
error grows by a factor of \sensTwSeRatio, and detection falls more for
not-direct pairs (\sensTwFirmDetNot to \sensTwFqDetNot) than for direct ones
(\sensTwFirmDetDir to \sensTwFqDetDir). Firm scaling divides by a median taken
over the whole window, after many of the actions. Taking it over the first
eight quarters instead gives \sensPreScaleAuroc (\sensPreScaleWT within
target).

\paragraph{Timing within the filing quarter.}
States are stamped at the start of a quarter, so the change at $h=1$ spans the
filing quarter and part of it can precede the filing: a bad January can
prompt an 8-K in March. Dropping $h=1$ removes that ambiguity. On $h=2,3,4$
the marginal test gives 0.612 against a restricted shift null of 0.537, an
excess of 0.076 against \excessShiftFirm on all four horizons (100
replicates), and 0.585 within target against 0.564. The $h=2$ change alone,
with flows measured as $s_{t+2}-s_t$ so that nothing of the filing quarter
enters, gives 0.570 against 0.520. Lagged cross-correlation computed on the
marginal test's panel, not the configuration selected for \cref{tab:methods},
moves from 0.705 to 0.692 within action, still 0.041 above its binding null
(40 replicates), and from 0.669 to 0.585 within target.

For survival, the marginal and expert-parent fits share one complete-case
sample, smaller than the headline
one because each row needs the parents' lags, and cover all targets, stocks
included. A pair is detected if any horizon clears the \v{S}id\'ak level for
four horizons, and survives if it stays detected given the expert parents at
its peak horizon. Detection falls, as it must when the horizon holding the
first effect is removed, from 46.8\% to 35.1\% of direct pairs and from
34.5\% to 17.0\% of not-direct ones, but survival does not: of the 73
not-direct pairs detected on $h=2,3,4$, 71\% survive, against 63\% of the 148
detected on all four (62\% and 69\% of direct pairs). On flow targets alone,
the population of \cref{tab:oracle-detect}, not-direct survival rises from
67\% (61 of 91) to 74\% (34 of 46); the table's 64\% comes from the
separate fits described at the start of \cref{app:oracle}, on their own sample. The results are not
driven by the horizon at which the order within a quarter is ambiguous.

\paragraph{Label error and confounding behind the not-direct detections.}
Of the \sensDetNotN not-direct pairs the marginal test detects on the
complete-case sample above, \sensDetNotVoted
carry any direct vote, and Dawid--Skene expects \sensDetNotDs of the \sensDetNotN
(\sensDetNotDsU) to be truly direct. Dawid--Skene is a sensitivity model, not
a measure of the graph's accuracy: it infers errors from the pattern of
disagreement, so a mistake all three annotators share is invisible to it.
Label errors consistent with that disagreement explain few of the
detections. For
confounding we compute, at each detected pair's strongest horizon, the
robustness value of \citet{cinelli2020making}: the share of the residual
variance of both the action and the outcome that a confounder must explain to
erase the estimate. Its median is \sensRvDir for both classes: a confounder
explaining 1.3\% of the residual variance of both the action and the outcome
would erase the median detection. With about
80{,}000 firm-quarters per cell the robustness value is close to
$|t|/\sqrt{\mathrm{df}}$, so any $|t|$ near 3 gives a value near 1\%;
\citet{cinelli2020making} read it against observed covariates, and so do we.
Log firm size explains a median 0.27\% of the action's residual variance and
0.05\% of the outcome's. Erasing the median detected not-direct pair takes a
confounder \sensKSizeNot times as strong as log firm size, \sensKActNot times
as strong as the firm's other actions and \sensKQuarterNot times as strong as
quarter shocks (\sensKSizeDir and \sensKActDir for direct pairs), and the
target's own change before the filing erases neither within 50 times its
strength. Relative to these observed benchmarks, erasing the median
association takes a confounder far stronger, in its partial associations
with both action and outcome, than any covariate we observe. By the same
measure, the not-direct detections are as robust as the direct ones the graph
accepts.

\section{Per-state tables}
\label{app:states}

\Cref{tab:persistence,tab:tails} give, per state, the persistence of
\cref{sec:persistence} and the tails of \cref{sec:tails}.

\begin{table}[h]
\centering
\caption{Within-firm persistence of the scaled levels. $\hat\rho$ is the median
per-firm OLS coefficient and IQR its interquartile range across firms;
$\hat\rho_{\mathrm{K}}$ is the median after Kendall's correction, with the
share of firms above one. A firm is dropped from a state's row if that state
is observed in fewer than 12 of its quarters.}
\label{tab:persistence}
\small
\begin{tabular}{@{}lcccccr@{}}
\toprule
State & $\hat\rho$ & IQR & $\hat\rho_{\mathrm{K}}$ & $\hat\rho_{\mathrm{K}}>1$ & ICC & Firms \\
\midrule
Equity & 0.93 & [0.84, 0.99] & 1.05 & 64\% & 0.78 & 2,901 \\
Noncurrent assets & 0.91 & [0.82, 0.97] & 1.03 & 59\% & 0.26 & 2,264 \\
Current assets & 0.87 & [0.74, 0.95] & 0.99 & 47\% & 0.37 & 2,901 \\
Current liabilities & 0.84 & [0.68, 0.94] & 0.96 & 42\% & 0.80 & 2,901 \\
Revenues & 0.83 & [0.58, 0.96] & 0.95 & 42\% & 0.70 & 2,741 \\
Operating income & 0.43 & [0.14, 0.69] & 0.51 & 8\% & 0.70 & 2,888 \\
Net income & 0.28 & [0.05, 0.57] & 0.34 & 4\% & 0.62 & 2,893 \\
Operating cash flow & 0.14 & [-0.09, 0.43] & 0.19 & 3\% & 0.64 & 2,901 \\
Investing cash flow & 0.05 & [-0.07, 0.25] & 0.09 & 1\% & 0.07 & 2,815 \\
Financing cash flow & -0.01 & [-0.12, 0.16] & 0.03 & 1\% & 0.38 & 2,890 \\
Other comprehensive income & -0.05 & [-0.21, 0.12] & -0.03 & 0\% & 0.02 & 2,339 \\
\bottomrule
\end{tabular}

\end{table}

\begin{table}[h]
\centering
\caption{Distribution of the scaled levels, as reported and after 1\%
winsorisation within quarter. Skew and the share of the sum of squares in the
top 1\% of rows are after winsorisation.}
\label{tab:tails}
\small
\begin{tabular}{@{}lrrrr@{}}
\toprule
 & \multicolumn{2}{c}{Excess kurtosis} & & Top 1\% \\
\cmidrule(lr){2-3}
State & raw & winsorised & Skew & SS share \\
\midrule
Noncurrent assets & 9,769 & 451 & 16.2 & 59\% \\
Financing cash flow & 77,082 & 86 & 7.6 & 65\% \\
Current assets & 10,009 & 76 & 5.6 & 27\% \\
Current liabilities & 9,495 & 68 & 7.9 & 76\% \\
Net income & 81,358 & 66 & -7.6 & 74\% \\
Equity & 4,043 & 60 & -7.3 & 73\% \\
Operating income & 83,688 & 55 & -6.9 & 69\% \\
Investing cash flow & 37,863 & 47 & -4.2 & 48\% \\
Operating cash flow & 47,968 & 38 & -5.4 & 57\% \\
Revenues & 15,270 & 19 & 3.3 & 28\% \\
Other comprehensive income & 50,847 & 18 & -0.7 & 37\% \\
\bottomrule
\end{tabular}

\end{table}

\end{document}